%% file: arxiv.tex
\PassOptionsToPackage{table}{xcolor}
\documentclass[10pt,twocolumn]{article}
\usepackage[letterpaper,margin=0.72in,columnsep=0.28in]{geometry}
\usepackage{arena-brand}
\usepackage{natbib}
\setcitestyle{numbers,square,comma}

\usepackage{amsmath,amssymb}
\usepackage{array}
\usepackage{booktabs}
\usepackage{graphicx}
\usepackage{listings}
\usepackage{multirow}
\usepackage{tabularx}
\usepackage{threeparttable}
\usepackage[breakable,skins,listings]{tcolorbox}
\usepackage{titlesec}
\usepackage{caption}
\usepackage{hyperref}
\usepackage[capitalize,nameinlink]{cleveref}

\titleformat{\section}
  {\ArenaHeadingFont\color{ArenaInk}\fontsize{16}{19}\selectfont}
  {\thesection.}{0.6em}{}
\titleformat{\subsection}
  {\ArenaHeadingFont\color{ArenaInk}\fontsize{13.5}{16}\selectfont}
  {\thesubsection.}{0.6em}{}
\titleformat{\subsubsection}
  {\ArenaHeadingFont\color{ArenaInk}\fontsize{12}{14.5}\selectfont\bfseries}
  {\thesubsubsection.}{0.6em}{}
\titlespacing*{\section}{0pt}{2.6ex plus 0.5ex minus 0.2ex}{1.0ex}
\titlespacing*{\subsection}{0pt}{2.0ex plus 0.4ex minus 0.2ex}{0.7ex}
\titlespacing*{\paragraph}{0pt}{1.25ex}{1em}
\lstdefinestyle{prompt}{
  basicstyle=\scriptsize\ttfamily\color{ArenaTextSecondary},
  breaklines=true,
  breakindent=0pt,
  columns=fullflexible,
  keepspaces=true,
  frame=single,
  rulecolor=\color{ArenaBorder},
  framerule=0.4pt,
  backgroundcolor=\color{ArenaPaper},
  aboveskip=6pt,
  belowskip=6pt,
  literate={–}{{\textendash}}1 {—}{{\textemdash}}1,
}

\title{Post-Training Frontier Text-to-Image Models by Composing Preference and Rubric Rewards}

\author{%
  Yuanhao Ban$^{1,2}$ \quad I-Hung Hsu$^{1}$ \quad Anastasios Angelopoulos$^{1}$\\
  Wei-Lin Chiang$^{1}$ \quad Ion Stoica$^{1}$ \quad Cho-Jui Hsieh$^{1,2}$
}
\newcommand{\affiliations}{%
  $^{1}$Arena Intelligence Inc. \qquad $^{2}$University of California, Los Angeles
}

\makeatletter
\newcommand{\ArenaTitleBlock}{%
  {\raggedright\setlength{\parindent}{0pt}%
    \hbox to \linewidth{\ArenaLogo[0.256in]\hfil
      \raisebox{3pt}{\ttfamily\fontsize{9}{11}\selectfont
        \color{ArenaTextSecondary}\today}}\par
    \vspace{8pt}
    {\color{ArenaInk}\rule{\linewidth}{1pt}}\par
    \vspace{16pt}
    {\ArenaTitleFont\fontsize{28}{33}\selectfont\color{ArenaInk}\@title\par}
    \vspace{12pt}
    {\fontsize{11.5}{14}\selectfont\@author\par}
    \vspace{6pt}
    {\fontsize{9.5}{11.5}\selectfont\color{ArenaTextSecondary}\affiliations\par}
  }
}
\makeatother

\renewenvironment{abstract}{%
  {\noindent\ArenaHeadingFont\large\bfseries Abstract\par}
  \vspace{8pt}\normalfont\normalsize
  \setlength{\parindent}{0pt}%
}{\par}

\hypersetup{
  colorlinks=true,
  linkcolor=ArenaInk,
  citecolor=ArenaTextSecondary,
  urlcolor=ArenaLink,
  pdftitle={Post-Training Frontier Text-to-Image Models by Composing Preference and Rubric Rewards},
  pdfauthor={Yuanhao Ban, I-Hung Hsu, Anastasios Angelopoulos, Wei-Lin Chiang, Ion Stoica, Cho-Jui Hsieh},
}

\begin{document}
\twocolumn[{
  \ArenaTitleBlock
  \vspace{12pt}
  \begin{tcolorbox}[enhanced,colback=ArenaPaper,colframe=ArenaBorderFaint,
    boxrule=0.75pt,arc=6pt,boxsep=0pt,
    left=14pt,right=14pt,top=10pt,bottom=10pt,
    before skip=0pt,after skip=14pt]
    \begin{abstract}
      \input{sec/0_abstract}
    \end{abstract}
  \end{tcolorbox}
}]

\input{sec/1_intro}
\input{sec/2_related}
\input{sec/3_method}
\input{sec/4_experiments}
\input{sec/5_limitations}
\input{sec/6_conclusion}
{\small
  \bibliographystyle{ieeenat_fullname}
  \bibliography{main}
}

\clearpage
\appendix
\input{sec/X_suppl}

\end{document}

%% file: sec/0_abstract.tex
Recent text-to-image generation models have achieved remarkable visual quality, but improving them through post-training remains challenging because no single reward signal captures the full range of human preference. In this work, we develop a simple and effective post-training recipe for open-domain text-to-image generation based on the composition of complementary reward signals. Our reward system consists of two main components: a preference reward, trained on large-scale human preference data using a Bradley–Terry objective to capture overall human aesthetic and perceptual preferences, and rubric-based rewards, which explicitly evaluate prompt faithfulness and other desirable properties while providing safeguards against reward hacking. A key challenge is how to combine these heterogeneous reward signals. We show that a naive weighted average leads to suboptimal optimization behavior, and propose a simple reward composition strategy that more effectively balances preference optimization with rubric satisfaction. In the Arena text-to-image leaderboard\footnote{\href{https://arena.ai/}{Arena leaderboard}}, our RL-trained Flux2dev achieves an Elo rating 69 points above the base model, and our post-trained Ideogram-4 surpasses every open-source model on the leaderboard, reaching an Elo of 1223.5.\footnote{Claims of state-of-the-art performance are based on the Arena leaderboard snapshot as of September 4, 2026.} Our results suggest that effective rewards for frontier generative-model training require broad coverage of user intent and robustness to exploitation under optimization. To support reproducible research, we release Arena-T2I-Training, a 1K subset of training data that
recovers some gains of full-scale training, providing a resource that we hope will
facilitate future work on post-training for text-to-image models. Project page: \href{https://banyuanhao.github.io/Post-Training-Frontier-Text-to-Image-Models-by-Composing-Preference-and-Rubric-Rewards/}{click here}.

%% file: sec/1_intro.tex
\section{Introduction}
\label{sec:intro}

Modern text-to-image (T2I) generation models have made rapid progress in visual quality, prompt understanding, and compositional ability. As base models become increasingly capable, an important question is how to further improve them with post-training. Several methods such as DiffusionNFT~\cite{zheng2026diffusionnft} and Flow-GRPO~\cite{liu2025flowgrpo}, have recently been developed for post-training T2I models, but their effectiveness critically depends on the rewards: open-domain image generation involves many aspects of quality that are difficult to capture with any single signal. Learned preference rewards such as PickScore~\cite{kirstain2023pickapic}, HPSv2/v3~\cite{wu2023hpsv2,ma2025hpsv3}, and ImageReward~\cite{xu2023imagereward} are trained on human preference data and provide broad supervision for what users like. Complementing them, recent work has developed more targeted rewards for specific capabilities such as spatial reasoning, text rendering, and compositional correctness~\cite{wei2025deepseek,long2026spatialreward}. However, naively combining these rewards does not necessarily improve performance, as poorly designed reward functions may fail to fully capture the desired behavior and can be hacked by the model.
\begin{table}[tb]
\centering
\footnotesize
\setlength{\tabcolsep}{5pt}
\renewcommand{\arraystretch}{0.92}
\caption{Arena scores on the Arena text-to-image leaderboard. Public scores are from the September 4, 2026 leaderboard snapshot.}
\label{tab:ledger-leaderboard}
\begin{tabular}{@{}lc@{}}
\toprule
Deployed model & Arena score \\
\midrule

\multicolumn{2}{@{}l}{\emph{FLUX.2-dev }} \\
base model
& $1132.8 \pm 14$ \\

Ours & $1202 \pm 13$ \\
\midrule
\multicolumn{2}{@{}l}{\emph{Ideogram-4}} \\

Base model (best open-source model) & $1204 \pm 4$ \\
Ours & $1224 \pm 13$ \\

\midrule
\multicolumn{2}{@{}l}{\emph{Public LMArena reference models}} \\
nano-banana-2
& $1261 \pm 5$ \\
seedream-5.0-pro
& $1258 \pm 4$ \\
nano-banana-pro
& $1232 \pm 5$ \\

Qwen-Image-2.0-Pro
& $1191 \pm 6$ \\
grok-imagine-image
& $1171 \pm 3$ \\
Cosmos3-Super-Text2Image (Agentic)
& $1168 \pm 8$ \\
FLUX.2-Max
& $1162 \pm 4$ \\
Hunyuan-image-3.0 \textnormal{(public)}
& $1151 \pm 3$ \\

\bottomrule
\end{tabular}
\end{table}

In this work, we develop a simple yet effective recipe for post-training modern open-domain T2I models by composing preference and rubric rewards. The preference term is a Bradley--Terry (BT)-based reward model trained on large-scale human preference data. While several existing reward models can serve this role, we find that scaling up the number of preference pairs yields further gains, and accordingly train ours on  $\sim 5$ million pairwise human preference votes collected from Arena. Preference supervision alone, however, is insufficient: it captures broad aspects of human preference such as aesthetics, yet remains largely blind to properties that are sparse or weakly represented in pairwise votes, such as faithfully including every requested object or satisfying prompt-specific constraints. Indeed, as shown in \cref{fig:goodhart-perdim}, when we directly optimize FLUX.2-dev~\cite{blackforestlabs2026flux2} against either our independently trained Arena reward model or the publicly available PickScore~\cite{kirstain2023pickapic}, the optimized reward continues to increase, while the actual quality quickly saturates and then degrades.

We address this limitation by complementing the broad preference signal with a set of open-ended, prompt-conditioned \emph{rubric rewards}, each designed to capture a distinct aspect of generation quality or failure mode. First, a \emph{faithfulness reward} explicitly evaluates whether the generated image contains the content and satisfies the requirements specified in the prompt~\cite{ban2026arenat2ihard}. Second, an intent-gated \emph{constraint reward} penalizes unsupported additions when literal adherence to the prompt is desired. Finally, to mitigate reward hacking, we introduce rubric-based \emph{hack detectors} that identify known undesirable behaviors emerging during optimization.

Another key challenge is how to combine these heterogeneous reward signals effectively. Our design consists of four components. First, we normalize each reward component to a common $[0,1]$ scale before aggregation~\cite{liu2026gdpo}, reducing sensitivity to differences in their native scales. Second, we develop hack detectors that identify exploits of the preference reward. Rather than treating their outputs as additional additive terms, we use them to suppress the positive preference reward associated with the detected exploit; this prevents the policy from gaining reward merely by learning to satisfy the detectors. This prevents the policy from gaining reward merely by learning to satisfy the detector. Furthermore, to reduce interference among competing reward signals, we introduce a gating mechanism that determines when individual rewards are active. Finally, we find that different reward configurations exhibit complementary strengths and failure modes. Therefore, we ensemble independently optimized policies directly in weight space by averaging their LoRA deltas~\cite{wortsman2022modelsoups}. Together, these design choices yield a robust reward ensemble that generalizes well across settings and substantially reduces the need for burdensome hyperparameter tuning.

Using this composed reward, we develop a post-training recipe that consistently improves already-strong T2I models under real-world human evaluation. We apply the same recipe to multiple frontier base models, including FLUX.2-dev and Ideogram-4, and observe substantial improvements over their corresponding base models. As shown in \cref{tab:ledger-leaderboard}, when evaluated through blind pairwise comparisons by real users on the Arena text-to-image leaderboard, our RL-trained FLUX.2-dev model achieves an Elo rating $69$ points higher than the base model. Applying the recipe to Ideogram-4 yields a model with an Elo score of $1223.5$, outperforming every open-source text-to-image model on the leaderboard. To facilitate reproducibility and further research, we will also release Arena-T2I-Training, a 1K subset of our training data and report the performance. We hope this can provide a lightweight setting for reproducing and studying our post-training recipe. These results suggest that large-scale preference learning and explicit rubric-based evaluation are not competing approaches to reward design; rather, their composition provides a scalable and effective recipe for improving modern text-to-image generation models through post-training.

%% file: sec/2_related.tex
\section{Related Work}
\label{sec:related}

\paragraph{RL for diffusion and flow models}
A growing body of work adapts reinforcement and preference optimization to diffusion and flow-based generative models. Diffusion-DPO~\cite{wallace2024diffusiondpo} adapts direct preference optimization~\cite{rafailov2023dpo} to diffusion likelihoods, avoiding an explicit reward model. Flow-GRPO~\cite{liu2025flowgrpo} brings GRPO's~\cite{shao2024deepseekmath} critic-free, group-relative optimization to flow matching through an ODE-to-SDE conversion. DiffusionNFT~\cite{zheng2026diffusionnft} instead performs policy improvement directly on the forward process, avoiding optimization through a reverse sampler. MixGRPO~\cite{li2025mixgrpo} studies mixed ODE/SDE rollout schedules. TempFlow-GRPO~\cite{he2025tempflowgrpo} introduces noise-aware timestep weighting.

\paragraph{Preference and quality rewards for T2I}
Text-to-image post-training uses several kinds of reward signals. Learned preference models such as ImageReward~\cite{xu2023imagereward}, PickScore~\cite{kirstain2023pickapic}, and HPSv2/v3~\cite{wu2023hpsv2,ma2025hpsv3} train scalar predictors from large-scale human comparisons of images, while UnifiedReward~\cite{wang2025unified} uses a unified VLM judge for both image understanding and generation quality. Other rewards target narrower notions of quality: perceptual scorers such as the LAION aesthetic predictor~\cite{schuhmann2022aesthetics} evaluate visual quality without explicitly measuring prompt adherence, while verifiable rewards measure specific capabilities programmatically, such as GenEval's detector-based object checks~\cite{ghosh2023geneval}, OCR-based text-rendering rewards~\cite{liu2025flowgrpo} or dependency-aware faithfulness rewards~\cite{ban2026arenat2ihard}.  We study two rewards from the learned-preference family: PickScore itself, used directly as a training objective, and an arena reward model trained on millions of pairwise human votes collected continuously from production models under real user prompts.

\paragraph{Rubric reward}
Rubric-based rewards decompose open-ended human preferences into explicit evaluation criteria, thereby extending reinforcement learning objectives from verifiable domains to more open-ended settings~\cite{gunjal2026rubrics}. OpenRubrics~\cite{liu2026openrubrics} derives rubrics by contrasting preferred and rejected responses, while AutoRule~\cite{wang2025autorule} uses chain-of-thought prompting on preference pairs to extract candidate evaluation rules. Autorubric-T2I~\cite{kao2026autorubric}  further learns rule weights and selects discriminative criteria using $\ell_1$-regularized logistic regression on image preference data, yielding an interpretable rubric-based reward model. Chasing the Tail~\cite{zhang2026chasing} iteratively sharpens rubrics to distinguish among top-performing responses. Auto-Rubric~\cite{tian2026auto} uses information-theoretic global compression to distill a compact, non-redundant set of reusable evaluation criteria from limited preference data.

%% file: sec/3_method.tex
\section{Method}
\label{sec:method}

We introduce the the preliminaries of online diffisuion RL in \cref{sec:m-nft}. Then We discuss our reward components and how we ensemble the rewards for RL training in \cref{sec:m-rubric} and \cref{sec:m-ensemble}, respectively.

\subsection{Preliminaries: Online Diffusion RL on the Forward Process}
\label{sec:m-nft}

We post-train the policy with
DiffusionNFT~\cite{zheng2026diffusionnft}, which applies reinforcement learning to the forward diffusion process.  For a  given rollout, let $v_\theta$ and
$v_{\theta_{\mathrm{old}}}$ denote the velocity prediction of the current policy and of the policy used to generate the rollout, respectively. DiffusionNFT forms a positive and implicit-negative
predictions
\begin{equation}
    v^{+}
    =
    (1-\beta)v_{\theta_{\mathrm{old}}}
    +
    \beta v_\theta,
    \qquad
    v^{-}
    =
    (1+\beta)v_{\theta_{\mathrm{old}}}
    -
    \beta v_\theta,
    \label{eq:nft-pm}
\end{equation}
and optimizes
\begin{equation}
    \mathcal{L}_{\mathrm{NFT}}
    =
    \tilde r_i
    \left\lVert v^{+} - v_{\mathrm{target}}\right\rVert_2^2
    +
    (1-\tilde r_i)
    \left\lVert v^{-} - v_{\mathrm{target}}\right\rVert_2^2,
    \label{eq:nft-loss}
\end{equation}
where $\tilde r_i\in[0,1]$ is a group-relative weight derived from the
reward of rollout $i$. We describe how $\tilde r_i$ is computed from our multi-axis reward in \cref{sec:m-ensemble}.

\subsection{Reward components}
\label{sec:m-rubric}

We use a learned human-preference reward to capture general visual quality, a presence-oriented faithfulness score to reward requested content, an absence-oriented constraint score to penalize unwanted additions, and a reward-hacking penalty to suppress observed reward-hacking behaviors.

\subsubsection{Human preference reward}
\label{sec:m-arena}
Our human preference reward is a Bradley--Terry preference model trained on
large-scale human pairwise votes. Given a \texttt{(prompt, image)} pair, it returns a
scalar preference logit. It captures holistic human
preference, including visual quality, composition, and aesthetics, but
does not explicitly monitor the prompt adherence. We adopt two human-preference reward models: the publicly available PickScore~\cite{kirstain2023pickapic} and an Arena reward model that we train on preference data collected through the Arena Leaderboard.

\paragraph{Arena reward model training}
\label{sec:data-collection}
Before training the model, we filter out careless voters and low-quality prompts to remove invalid votes. The final dataset spans
2025-04-16 to 2026-06-07 with a total of around 5 million vote pairs from $100+$ T2I models.
Vote-record schema, logging, and corpus-scale details
are in \cref{sec:suppl-rmdetails}. We use Qwen3.6-27B as the backbone and add a single linear head that reads the hidden state of a trailing
\texttt{<|cls|>} token and outputs a scalar Bradley--Terry logit.
We independently perturb image brightness, contrast, and
saturation using multiplicative color jitter, and add pixel-wise Gaussian
noise to improve robustness to low-level image statistics. We train for a single pass over the corpus
with a batch size of 1024. Additional optimization hyperparameters details are provided in
\cref{sec:suppl-rmdetails}.

\subsubsection{Rubric faithfulness reward}
\label{sec:m-faith}

The faithfulness reward measures whether the content requested by the
prompt is correctly realized in the generated image.
Several algorithms for this have been proposed in the literature to decompose each prompt into a dependency-aware set of yes/no questions~\cite{hu2023tifa, cho2024dsg, ban2026arenat2ihard}. Here we adopt the same methodology from Arena-T2I-Hard~\cite{ban2026arenat2ihard} which uses an LLM to decompose the prompt into a tree-structured checklist offline.
A VLM judge answers the
applicable questions, and the faithfulness reward is defined as the
fraction answered positively. This signal is presence-oriented: it evaluates whether requested objects, attributes, relations, and styles are present and correctly realized. In practice, we use \texttt{gemini-3-pro-preview} to decompose our training prompts, yielding $182$k questions over $10$k prompts.
We use Qwen3.6-27B as our VLM judge. Prompt
templates, statistics, and a worked example are provided in
\cref{sec:suppl-faithq}.

\subsubsection{Rubric constraint reward}
\label{sec:m-gate}

Furthermore, we introduce a complementary absence-oriented constraint reward that discourages content inconsistent with  the user's intent. Using the same VLM-based checklist framework as the faithfulness reward, we construct questions that check whether the generation satisfies constraints on undesirable or extraneous content. These constraints include explicit negative instructions, restrictions on background, style, or additional objects, as well as implicit requirements inferred from the user's intent. For example, given a prompt requesting a minimalist image, introducing unnecessary or unrequested elements may violate the intended simplicity even though no such element is explicitly prohibited. In practice, the constraint checklist for each prompt is generated offline by a single call to  \texttt{gemini-3.1-pro-preview}, yielding $26$k constraint questions over the $10$k training prompts. Each question is phrased so that a ``yes'' answer indicates compliance,  and the constraint reward is the fraction of questions answered ``yes'' by the same Qwen3.6-27B judge during training. Additional details are provided in \cref{sec:suppl-constrq}.

\subsubsection{Rubric anti-reward-hack reward}
\label{sec:m-safeguard}

To prevent the policy from exploiting loopholes in the reward model, we introduce rubric-based anti-reward-hack rewards that target known reward-hacking behaviors. Depending on the nature of the failure mode, we consider two general classes of anti-reward-hack rewards:

\paragraph{Pointwise anti-reward-hack rewards}
When a failure can be evaluated from a single generated image, we use a pointwise anti-reward-hack reward that directly checks  whether the output exhibits the corresponding undesirable behavior. This form is appropriate for criteria that admit an approximately absolute notion of correctness and therefore do not require comparison with a reference generation. For example, we observe that some reward models can encourage the generation of unrequested garbled text. As malformed text can be identified directly from a single image, this failure is naturally captured by a pointwise anti-reward-hack reward.

\paragraph{Reference-based anti-reward-hack rewards}
For failures defined by an undesirable drift of the policy relative to the base model, we instead use reference-based anti-reward-hack rewards that compare the policy output against a generation from the frozen base model for the same prompt. Such comparisons make the direction of the policy shift explicit and assess whether the policy output exhibits stronger evidence of the undesirable behavior than the reference. For example, we observe a tendency toward excessive photorealism even for prompts that explicitly request a cartoon style. However, from a single image alone, it can be difficult to determine whether the degree of photorealism is excessive. By comparing the policy output with a frozen-base generation for the same prompt, the detector can more reliably identify whether the policy has shifted further toward photorealism. A reference-based anti-reward-hack reward is therefore better suited to capturing this type of relative failure. To mitigate position bias, each pair is judged in both presentation orders. All detector verdicts are produced by the same Qwen3.6-27B judge used for the checklist rewards. The verbatim detector questions and their arming conditions are listed in \cref{sec:suppl-hackdetectors}.

\subsection{Reward Ensembling}
\label{sec:m-ensemble}

The reward signals described above operate are heterogeneous: they  are defined on different native scales, and each captures a distinct aspect of generation quality.  Naively summing them can lead to interference among reward components, yielding an aggregate reward that is unstable and difficult to optimize. To address this issue, we introduce three key mechanisms: reward-decoupled normalization, prompt-conditioned gating, and a veto-based anti-reward-hack prevention.

\subsubsection{Group reward-decoupled normalization}

Following the GDPO-style reward-decoupled normalization scheme~\cite{liu2026gdpo}, we normalize each reward axis independently:
\begin{equation}
z_i^{(a)}
=
\frac{r_i^{(a)}-\mu^{(a)}}
{\sigma^{(a)}+\epsilon},
\qquad
a \in \mathcal{A}
=
\{\mathrm{pref}, \mathrm{faith}, \mathrm{constr}\},
\label{eq:axis-norm}
\end{equation}
where $r_i^{(a)}$ denotes the raw reward of sample $i$ on axis $a$, and $\mu^{(a)}$ and $\sigma^{(a)}$ denote the corresponding mean and standard deviation computed over the current batch.
Normalizing each axis
independently, rather than the summed reward, prevents differences in reward
scale from causing any single component to dominate the combined optimization
objective, thereby improving training stability.

\subsubsection{Gating mechanism}

Not every reward signal is meaningful for every prompt. In particular, constraint-based rewards may be inappropriate for prompts that allow substantial creative freedom. Consider, for example, the prompt ``create an image of a country road in an Impressionist style.'' Adding elements such as a woman or flowers is a legitimate creative choice rather than a violation of the user's intent, so applying the constraint reward indiscriminately would penalize valid generations and unnecessarily restrict the model's generation space. The anti-reward-hack rewards are prompt-dependent in the same way. For example, a tendency toward photorealism is not undesirable when the prompt explicitly requests a photorealistic image, yet the same tendency is a failure when the prompt requests a cartoon style. We therefore introduce prompt-conditioned applicability gates for both the constraint reward and the anti-reward-hack rewards.

For the constraint reward, we infer a user-intent label for each prompt using an offline LLM classifier, which assigns one of \emph{strict}, \emph{neutral}, or \emph{open}. The constraint reward is activated only for prompts classified as \emph{strict}. We define
\begin{equation}
g_i^{(a)}
=
\begin{cases}
1,
& a \in \{\mathrm{pref}, \mathrm{faith}\},\\
\mathbf{1}
[y_i^{\mathrm{intent}}=\mathrm{strict}],
& a=\mathrm{constr}.
\end{cases}
\label{eq:reward-gating}
\end{equation}

In practice, we use \texttt{gemini-3.1-pro-preview} to classify the
training prompts. The labeling criteria and examples for each category are provided in \cref{sec:suppl-constrq}. To validate the gate labels, two authors
blindly labeled $400$ evaluation prompts. Human--classifier agreement is $77.8\%$ across the three-way classification, increasing to $89.5\%$ for the binary strict-versus-nonstrict decision actually used by the gate. Full protocol and the confusion matrix are in \cref{sec:suppl-gateaudit}

For each anti-reward-hack reward $h \in \mathcal{H}$, where $\mathcal{H}$ denotes the set of anti-reward-hack rewards, we additionally use an offline LLM classifier to decide whether the corresponding failure mode is applicable to the prompt:
\begin{equation}
g_i^{(h)}
=
\mathbf{1}
[y_{i,h}^{\mathrm{app}}=\mathrm{applicable}],
\qquad h \in \mathcal{H}.
\label{eq:hack-gating}
\end{equation}
The preference and faithfulness rewards are thus always active, the constraint reward applies only to strict-intent prompts, and each anti-reward-hack reward is considered only when its corresponding failure mode conflicts with the user's intent. This prompt-conditioned gating prevents irrelevant reward signals from introducing conflicting optimization pressure. In practice, we use \texttt{gemini-3.1-pro-preview} to label whether the corresponding failure mode would be applicable for each training prompt, with a single call per prompt covering all failure modes. The detector applicability criteria are listed in \cref{sec:suppl-hackdetectors}.

\subsubsection{Veto-based anti-reward-hack prevention}
\label{sec:m-veto}

Rather than treating anti-reward-hack rewards as additional additive reward terms, we use them as \emph{vetoes} on the preference reward. When an applicable anti-reward-hack reward identifies a known exploit, it suppresses any positive preference reward causing the issue; otherwise, the preference signal remains unchanged. Crucially, avoiding a reward-hacking behavior earns no additional reward. This one-sided design prevents the anti-reward-hack rewards from introducing new optimization targets that could themselves be exploited.

Let $r_i^{(h)} \in \{0,1\}$ indicate whether anti-reward-hack reward $h$ detecets the corresponding failure for sample $i$. We define the overall veto indicator as
\begin{equation}
v_i
=
\max_{h \in \mathcal{H}}
g_i^{(h)} r_i^{(h)}.
\label{eq:veto-indicator}
\end{equation}
Thus, $v_i=1$ if at least one applicable anti-reward-hack reward identifies a violation, and $v_i=0$ otherwise. We then define the safeguarded preference reward as
\begin{equation}
\bar z_i^{(\mathrm{pref})}
=
z_i^{(\mathrm{pref})}
-
v_i
\left[z_i^{(\mathrm{pref})}\right]_+,
\label{eq:veto}
\end{equation}
where $[x]_+=\max(x,0)$. When $v_i=1$, any positive preference reward is vetoed, while a non-positive preference reward remains unchanged. When no applicable violation is detected, $v_i=0$ and the original preference reward is preserved.

\subsubsection{RL objective}

After applying normalization, prompt-conditioned gating, and the anti-reward-hack veto, we combine the remaining reward signals as
\begin{equation}
R_i
=
w_{\mathrm{pref}}\,\bar z_i^{(\mathrm{pref})}
+
w_{\mathrm{faith}}\,z_i^{(\mathrm{faith})}
+
w_{\mathrm{constr}}\,g_i^{(\mathrm{constr})}z_i^{(\mathrm{constr})}.
\label{eq:reward-ensemble}
\end{equation}
Here, $w_a$ controls the relative contribution of reward axis $a$. In all experiments, we simply set $w_a=1$ for every reward axis. This untuned equal weighting performs consistently across both choices of preference reward model and both base models.

Finally, for the $K$ rollouts associated with the same prompt, we convert the combined reward into the group-relative weight used by DiffusionNFT:
\begin{equation}
A_i
=
\frac{R_i-\mu_{\mathrm{group}}}
{\sigma_{\mathrm{batch}}},
\qquad
\tilde r_i
=
\frac{1}{2}
\operatorname{clip}
\left(
\frac{A_i}{A_{\max}},
-1,
1
\right)
+
\frac{1}{2},
\label{eq:advantage}
\end{equation}
where $\mu_{\mathrm{group}}$ denotes the mean reward over the $K$ rollouts associated with the same prompt, $\sigma_{\mathrm{batch}}$ denotes the standard deviation of rewards over the current batch, and $A_{\max}=5$.

\subsubsection{Weight ensembling}
\label{sec:m-crosssoup}

Different reward configurations can steer the policy toward complementary solutions, with different runs specializing in distinct aspects of generation quality. We find that this complementarity also persists at the parameter level, motivating us to ensemble independently optimized policies directly in \emph{weight space}. In particular, we combine two runs trained with different reward configurations: one without anti-reward-hack rewards and one with anti-reward-hack prevention. Given their LoRA parameter deltas $\Delta\theta_1$ and $\Delta\theta_2$, we construct the ensembled adapter using simple equal-weight averaging:
\begin{equation}
\Delta\theta_{\mathrm{ens}}
=
\frac{1}{2}\Delta\theta_1
+
\frac{1}{2}\Delta\theta_2.
\end{equation}
This 1:1 weight soup requires no additional training or optimization and provides a simple way to combine complementary behaviors learned under different reward configurations. We use equal weighting.
Further analysis of the complementarity between runs, together with detailed  soup recipes, is provided in \cref{sec:suppl-soups}.

%% file: sec/4_experiments.tex
\section{Experiments}
\label{sec:experiments}

This section shows our experiments on post-training T2I models. \cref{sec:setup} presents the post-training setup and evaluation protocol. In \cref{sec:main-results}, we introduces the main results of our methods. \cref{sec:ablation} studies the key designs of our reward ensemble mechanism.  In \cref{sec:ideo-live}, we generalize our method to another base model. Finally, \cref{sec:leaderboard} reports the Elo score of our post-trained model on the Arena Leaderboard.

\subsection{RL settings}
\label{sec:setup}

\paragraph{Training and evaluation prompts} We sample training and validation prompts from the distribution of real user prompts in Arena leaderboard. We found RL is more sensitive to data quality than the RM training. So we use a stricter system prompt with \texttt{gemini-3.5-flash} to filter the prompt pool more aggressively, removing image-editing requests, under-specified one-line prompts, prompts centered on niche named entities, and non-English prompts. This process yields 10k training prompts and 1k validation prompts. More details on the prompt filter are provided in \cref{sec:suppl-promptfilter}, together with an early controlled comparison in which training on the unfiltered prompt pool degrades substantially faster under the same reward.

\begin{figure*}[t]
  \centering
  \includegraphics[width=\textwidth]{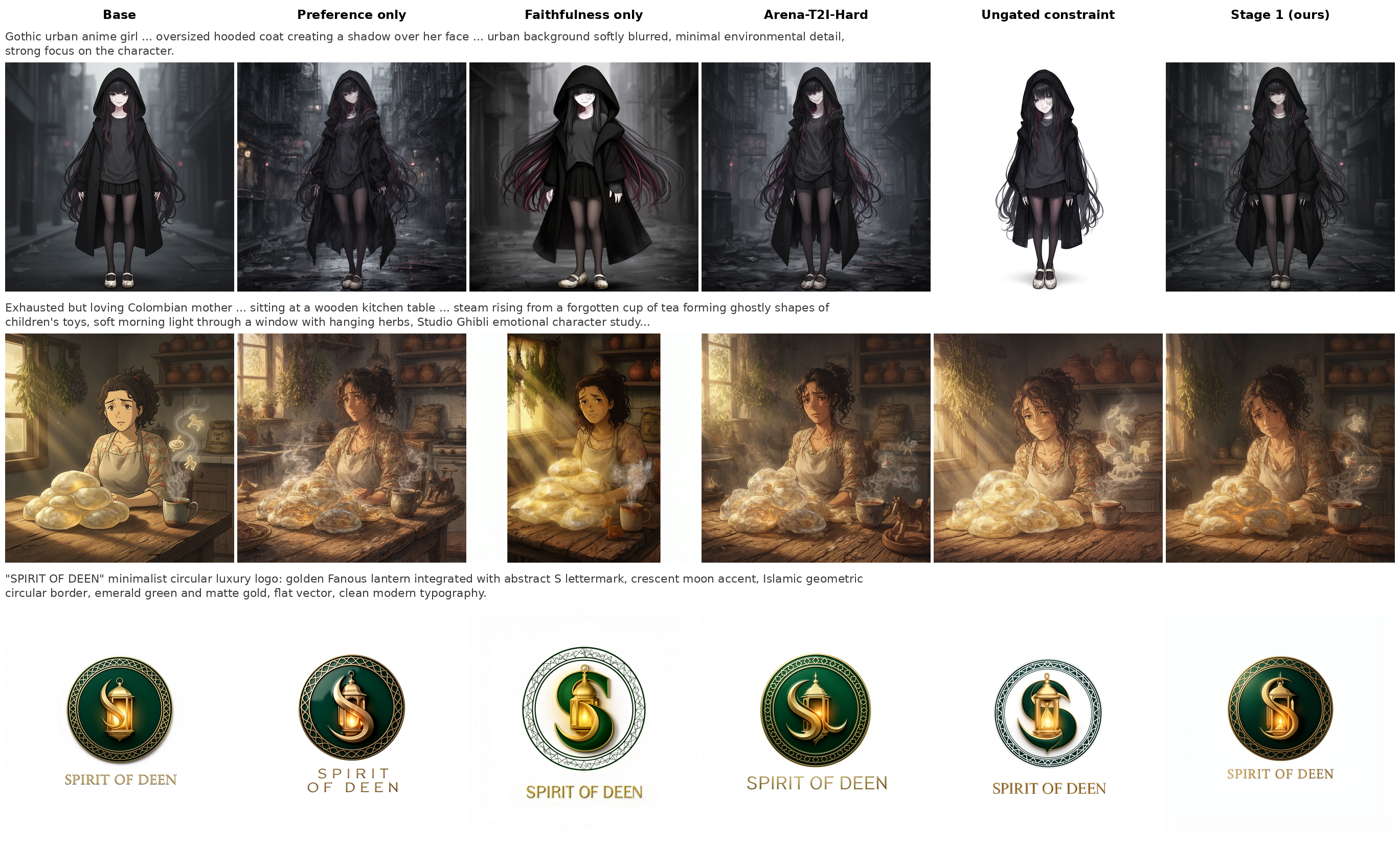}
  \caption{Visualization. Preference-only training drifts
    toward generic embellishment
    and ignores requested content; faithfulness-only training
    washes out detail and drops requested style and scenery;
    Arena-T2I-Hard recovers the requested
    elements but keeps unrequested decoration; the ungated constraint flattens detail even where the prompt
    invites it; the intent-gated Stage~1 keeps the constraint pressure
    only where the prompt closes the spec, yielding compliant
    renders without an obvious quality tax.}
  \label{fig:qual-rewarddesign}
\end{figure*}

\paragraph{Base models.}
We adopt two frontier text-to-image foundation models as our base model: FLUX.2 [dev]~\cite{blackforestlabs2026flux2} and Ideogram 4~\cite{ideogram2025ideogram}. FLUX.2 [dev] is a 32B-parameter rectified-flow transformer with a hybrid multimodal DiT architecture, comprising 8 double-stream blocks that jointly model text and image features while maintaining modality-specific representations, followed by 48 single-stream blocks operating on the concatenated token sequence. It uses Mistral Small 3.1~\cite{mistral2025small31} for text conditioning. In contrast, Ideogram 4 is a 9.3B-parameter flow-matching model built on a fully single-stream, 34-layer DiT, where text and image tokens are jointly processed throughout the network. It uses Qwen3-VL-8B-Instruct~\cite{bai2025qwen3} as its text encoder and demonstrates strong capabilities in prompt following, typography, and text rendering. Currently in Arena Leaderboard, Ideogram 4 is the open source Text-to-image generative model \#1.

\paragraph{Training configuration}
\label{sec:trainconfig}  We use DiffusionNFT~\cite{zheng2026diffusionnft} as our optimizer.
For Flux.2[dev], each step samples $G{=}48$ prompt groups and generates
$K{=}24$ rollouts per prompt using 10 sampling steps at $512^2$
resolution with CFG $4.0$. Ideogram 4 uses the same group layout and rollout budget, yet sampled with a constant guidance weight of $7.0$. Both models are fine-tuned with LoRA adapters on frozen base weights. Additional optimization hyperparameters are provided in
\cref{sec:suppl-hparams}.

\paragraph{Evaluation metrics}
To evaluate post-trained T2I models, we follow the MMRBv2~\cite{hu2026multimodal} evaluation protocol and use \texttt{gemini-3.5-flash} with the 1k Arena validation set to compare each trained checkpoint against the frozen base model. We report the pairwise win rate over the base model. For every run, we evaluate checkpoints ${30, 60, 90, 120, 150}$ on the $1$k Arena validation set and report the best-performing checkpoint. To test transferability, we also report results using another LLM judge\texttt{gpt5.4}. Rendering settings, the judging rubric, position debiasing and aggregation, and the full validation/held-out protocol are detailed in \cref{sec:suppl-eval}.
During inference, FLUX.2 generates images at a resolution of $1024^2$ using 50 sampling steps and a CFG scale of $4.0$, matching the default Hugging Face configuration. Ideogram 4 uses the official \emph{default\_quality\_48} configuration released in its repository, which performs 48 steps and a scheduled CFG weight of 7 for the first 45 steps and 3 for the final three polishing steps at a resolution of $1024^2$. Before generation, each evaluation prompt is expanded by \texttt{gemini-3.5-flash} using the model-specific official prompt-rewriting message. The upsampling message for FLUX.2 and the Magic Prompt message for Ideogram 4 are in \cref{sec:suppl-systemprompt}.

\paragraph{Baselines}
\emph{DiffusionNFT.} We follow the multi-reward recipe released with DiffusionNFT~\cite{zheng2026diffusionnft}. It optimizes a weighted combination of rewards over five stages of 30 steps each. Stages~1, 3, and~5 optimize PickScore~\cite{kirstain2023pickapic}${}+{}$HPSv2~\cite{wu2023hpsv2}${}+{}$CLIPScore~\cite{hessel2021clipscore} on the 10k Arena prompt set; stage~2 additionally incorporates a GenEval reward on GenEval prompts; and stage~4 additionally incorporates an OCR reward on OCR prompts. The GenEval and OCR training prompts are taken from the official DiffusionNFT GitHub repository. \emph{Arena-T2I-Hard.} We also include Arena-T2I-Hard as a natural baseline, which uses only a faithfulness reward and a preference reward. We reimplement it and train it on our 10k Arena training prompts using the same faithfulness questions as in our method.

\paragraph{Implementation details}
Our method consists of three stages. In Stage~1, we combine the preference, faithfulness, and intend-gated constraint rewards and train an initial post-trained model. In Stage~2, we analyze the reward-hacking failure modes exhibited by the best Stage~1 checkpoint, design corresponding anti-reward-hack rewards, and train a second model with these additional signals. In Stage~3, we combine the best checkpoints from Stages~1 and~2 through weight-space model souping to obtain the final model.

\paragraph{Open-sourced plan} To facilitate reproducibility, we plan to release a xxx subset of our training prompts. We retrain Stages~1 and~2 on this subset using PickScore as the preference reward, and report the resulting evaluation performance. This release is intended to enable researchers to independently reproduce our post-training pipeline.

\subsection{Main results}
\label{sec:main-results}
\cref{tab:all-results} summarizes our main results on the 1k Arena validation prompts. Our Stage~1 model, which uses the preference reward, faithfulness rewards and the intent-gated constraint reward, raises the win rate to $0.642$ with Arena RM and $0.581$ with PickScore. The improvement also transfers to the independent GPT-5.4 judge, where the same recipe reaches $0.561$ and $0.611$, respectively. Relative to Arena-T2I-Hard which lacks for gated-constrain reward, this corresponds to gains of $7.1$ and $4.9$ percentage points under Gemini-3.5-flash.

\begin{figure}[t]
  \centering
  \includegraphics[width=\linewidth]{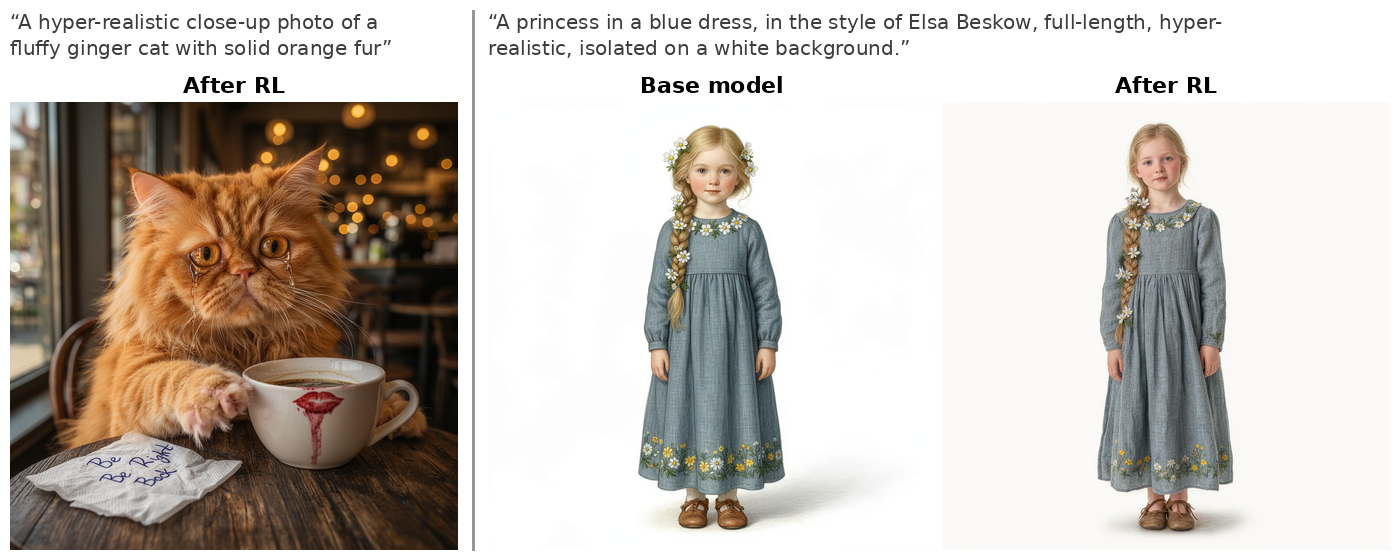}
  \caption{Two recurring failure modes.
  Left: the prompt requests no text, yet the RL policy introduces
  garbled strings on the napkin. Right: the prompt requests
  Elsa Beskow's storybook illustration style; the base model preserves
  the requested style, whereas the RL policy drifts toward photorealism.}
  \label{fig:hack-pairs}
\end{figure}

\paragraph{Anti-reward-hack reward} As shown in \cref{fig:hack-pairs}, we audit $205$
failure cases of the best Stage~1 checkpoint and identify two recurring \emph{reward-hack} modes: (i) garbled or unrequested rendered text,
and (ii) photo-style drift, where the policy replaces an explicitly
requested non-photographic style with cinematic photorealism. We therefore introduce targeted rubric-based detectors that
identify these exploits during training and prevent them from being
reinforced. Combining the new anti-reward-hack reward, our Stage~2 checkpoint is comparable to the previous one.

\paragraph{Weight ensembling} We continues to merge the model weight of our best Stage~1 checkpoint and the Stage~2 checkpoint. It is motivated by the fact that the two parent runs exhibit complementary behavior despite achieving similar aggregate win rates. On the 1k validation prompts, Stage~1 checkpoint has $204$ clear losses against the base model, while the gated-veto run has $234$, with only $92$ losses shared between them. Their wins also concentrate in different categories: the no-hack-prevention Stage~1 run performs particularly well on composition ($65\%$) and style ($16\%$), whereas the hack-prevention Stage~2 run achieves relatively more wins through faithfulness ($16\%$ vs.\ $9\%$) and clean text rendering ($9\%$ vs.\ $3\%$). As shown in \cref{tab:all-results}, the resulting model improves the Gemini win rate to $0.6598$ with Arena RM and $0.6256$ with PickScore, outperforming both parent models. Per-prompt analysis further shows that the soup retains complementary strengths from both parents: Stage~3 wins on $0.560$ and $0.397$ of the Stage~1's and Stage~2's exclusive loss sets, respectively, and still achieves a $0.304$ win rate on prompts where both parents lose. These results support the effectiveness of our weight ensembling mechanism.

\paragraph{Open-sourced results} On our open-sourced 1K datasets, Stage 1 and Stage 3 with PickScore achieve Gemini win rates of 0.566 and 0.615, respectively, approaching the performance obtained with the full 10K dataset. With the dataset and the publicly available PickScore model, anyone can reproduce our results. We hope these datasets can serve as a useful resource for the research community.

\paragraph{Qualitative results}
\cref{fig:qual-rewarddesign} qualitatively illustrates the same progression observed in the quantitative results. First optimizing a single preference reward leaves visible failure modes: arena-only training tends to drift toward generic embellishment and sometimes misses requested content. Arena-T2I-Hard combines the preference reward with a faithfulness reward and better recovers the requested elements. However, it might encourage the model to over-satisfy the prompt by inserting extra objects that make the required concepts easier to recognize, which can make the rendered image cluttered and less preferred. As shown in the second row of \cref{fig:qual-rewarddesign}, the prompt asks for children's toys to appear only as ghostly shapes formed by steam rising from a forgotten cup of tea, but the Arena-T2I-Hard model additionally materializes a solid wooden toy horse on the table. Applying the constraint reward uniformly removes some of these additions, but can also over-correct. As seen in the first row of \cref{fig:qual-rewarddesign}, while the ungated constraint reward removes undesirable decorations, it also suppresses benign background details, collapsing the background into a flat white field. In contrast, in our Stage~1 run, we apply constraints only when the user calls for strict adherence, producing images that better preserve the requested content and style while avoiding unnecessary additions. These examples qualitatively mirror the improvements observed in the quantitative results.

\begin{table}[t]
\centering
\scriptsize
\setlength{\tabcolsep}{5pt}
\renewcommand{\arraystretch}{1.05}
\caption{All pairwise win rates vs.\ the frozen FLUX.2 base. Each cell reports the run's best checkpoint among ${30, 60, 90, 120, 150}$,
selected on the 1k Arena validation rewrite prompts. Please refer to \cref{tab:ledger-cis} for $95\%$ percentile-bootstrap confidence intervals for
every cell.}
\label{tab:all-results}
\resizebox{\columnwidth}{!}{%
\begin{tabular}{@{}l cc cc@{}}
\toprule
& \multicolumn{2}{c}{Gemini-3.5-flash}
& \multicolumn{2}{c}{GPT-5.4} \\
\cmidrule(lr){2-3}\cmidrule(l){4-5}
Id & Arena RM & PickScore & Arena RM & PickScore \\
\midrule

\multicolumn{5}{@{}l}{\textit{Baseline}} \\
DiffusionNFT
& .290 & .290 & .396 & .396 \\
faithfulness only
& .366 & .366 & .367 & .367 \\
Arena-T2I-Hard
& .589 & .577 & .557 & .537 \\
preference only
& .507 & .461 & .417 & .517 \\
\midrule
\multicolumn{5}{@{}l}{\textit{Ours}} \\
Stage~1 (w/o anti-hack)
& .642 & .581 & .561 & .611 \\
Stage~2 (w/ anti-hack)
& .635 & .571 & .567 & .585 \\
Stage~3 (Stage~1, Stage~2 soup)
& .660 & .626 & .572 & .623 \\

\midrule
\multicolumn{5}{@{}l}{\textit{Ours Open-sourced}} \\
Stage~1 (w/o anti-hack)
& --- & .566 & --- & .555 \\
Stage~2 (w/ anti-hack)
& --- & .584 & --- & .510 \\
Stage~3 (Stage~1, Stage~2 soup)
& --- & .615 & --- & .604 \\

\bottomrule
\end{tabular}
}
\end{table}

\subsection{Ablation study}
\label{sec:ablation-study}

We next ablate the four design choices underlying our final reward-composition recipe. \cref{tab:ablation} summarizes these controlled comparisons.

\paragraph{Constraint gate}
\label{sec:ablation}
We first isolate the effect of intent gating for the constraint reward. We compare an ungated variant, which applies the constraint reward to all prompts, against our intent-gated Stage~1 run. As shown in \cref{tab:ablation}, intent gating increases the Gemini win rate from $0.6046$ to $0.6416$ with Arena RM and from $0.5200$ to $0.5810$ with PickScore. The improvement across both preference rewards and both judges supports the use of intent-gated rather than uniform constraint application.

\paragraph{Anti-reward-hack applicability gating}
We next isolate the applicability gate for the anti-reward-hack rewards. We compare an ungated variant, in which detector vetoes are applied without checking prompt-level applicability, against the gated Stage~2 run. As shown in \cref{tab:ablation}, the effect of the gate is reward-dependent. With Arena RM, applicability gating raises the Gemini and Gpt5.4 win rate by $0.1086$ and $0.0940$, respectively.

\paragraph{Anti-reward-hack one-side veto mechanism}
We then study how anti-reward-hack detector signals should enter the training objective. To this end, we compare two formulations: treating detector compliance as an additive reward axis or using detector violations as a one-sided veto which is excatly Stage~2 run. We set $w=1$ for the additional additive anti-reward-hack reward. As shown in \cref{tab:ablation}, The one-sided veto run substantially outperforms the additive formulation. Under Gemini, the win rate increases from $0.5260$ to $0.6346$ with Arena RM and from $0.4194$ to $0.5710$ with PickScore. These results favor the one-sided veto over incorporating detector compliance as an additional positive reward.

\paragraph{Model ensembling}
Finally, we ablate the source models for weight ensembling. We compare averaging the best two checkpoints from the same Stage~2 run against averaging models obtained from two different reward recipes: Stage~1 and Stage~2. Both variants use weight souping~\cite{wortsman2022modelsoups} over LoRA deltas without additional optimization. Under Gemini, same-run souping achieves win rates of $0.6435$ with Arena RM and $0.5610$ with PickScore, whereas cross-recipe souping improves these to $0.6598$ and $0.6256$.  The consistent advantage of cross-recipe over same-run souping indicates that the ensembling gain is primarily associated with combining models trained under different reward recipes rather than checkpoint averaging within a single run. Exact soup recipes are provided in \cref{sec:suppl-soups}.

\begin{figure}[t]
  \centering
  \includegraphics[width=0.95\linewidth]{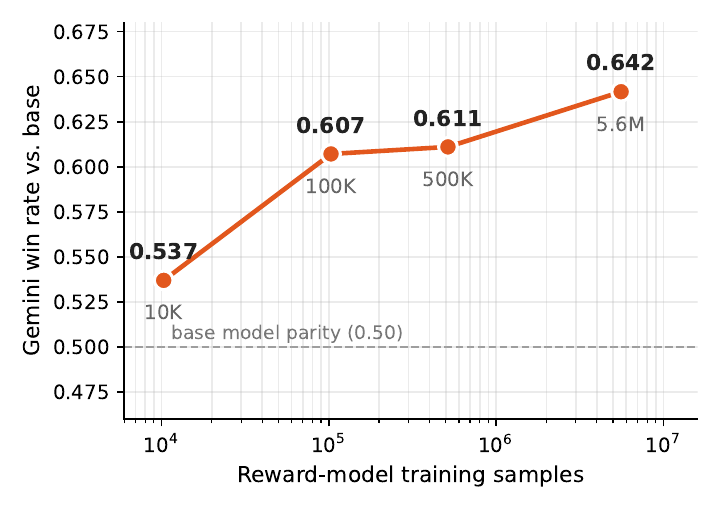}
  \caption{RL outcome scales with the preference reward model's training data. The x-axis shows the number of human preference pairs used to train the reward model, and the y-axis shows the MMRBv2 win rate of the resulting Stage~1 RL model against the frozen base.}
  \label{fig:rm-scaling}
\end{figure}

\paragraph{Reward-model data scaling Law}
We investigate how the amount of preference data used to train the reward model affects downstream RL performance.
Specifically, we use intermediate checkpoints of our Arena RM trained on ${\sim}10$K, ${\sim}100$K, and ${\sim}500$K pairwise human preference votes, apply the same Stage~1 RL recipe to each checkpoint, and evaluate the resulting models using MMRBv2 win rate against the frozen base model.
As shown in \cref{fig:rm-scaling}, RL performance improves as the amount of RM training data increases: the win rate rises from $0.537$ with 10K preference pairs to $0.607$ with 100K and $0.611$ with 500K, while the full ${\sim}5.6$M-pair RM achieves $0.642$.
These results indicate that scaling human preference data for reward-model training translates into consistent gains in downstream RL performance.

\begin{table}[htb]
  \centering
  \scriptsize
  \setlength{\tabcolsep}{2pt}
  \renewcommand{\arraystretch}{0.92}
  \caption{Ablation study on the ensembling mechanisms. Win rate vs.\ the base model on the 1k Arena
  validation prompts; each cell is the best checkpoint among
  $\{30,60,90,120,150\}$, selected under the Gemini judge.}
  \label{tab:ablation}
  \resizebox{\columnwidth}{!}{%
  \begin{tabular}{@{}llcccc@{}}
    \toprule
    & & \multicolumn{2}{c}{Gemini-3.5-flash} & \multicolumn{2}{c}{GPT-5.4} \\
    \cmidrule(lr){3-4}\cmidrule(l){5-6}
    Mechanism & Variant & Arena RM & PickScore & Arena RM & PickScore \\
    \midrule
    \multirow{2}{*}{Constrain Gate}
      & ungated & 0.6046 & 0.5200 & 0.5555 & 0.4985 \\
      & intent-gated & \textbf{0.6416} & \textbf{0.5810} & \textbf{0.5615} & \textbf{0.6105} \\
    \midrule
    \multirow{3}{*}{Hack applicability gate}
      & ungated & 0.5591 & 0.5695 & 0.5000 & 0.5515 \\
      & applicable-gated & \textbf{0.6346} & \textbf{0.5710} & \textbf{0.5670} & \textbf{0.5845} \\
    \midrule
    \multirow{3}{*}{Hack signal integration}
      & additive & 0.5260 & 0.4194 & 0.4730 & 0.4105 \\
      & one-sided veto & \textbf{0.6346} & \textbf{0.5710} & \textbf{0.5670} & \textbf{0.5845} \\
    \midrule
    \multirow{2}{*}{Weight ensembling}
      & same-recipe & 0.6435 & 0.5610 & 0.5465 & 0.5675 \\
      & cross-recipe & \textbf{0.6598} & \textbf{0.6256} & \textbf{0.5720} & \textbf{0.6225} \\
    \bottomrule
  \end{tabular}
  }
\end{table}

\subsection{A second base model: Ideogram-4}
\label{sec:ideo-live}

\begin{table}[t]
\centering
\scriptsize
\setlength{\tabcolsep}{5pt}
\renewcommand{\arraystretch}{1.05}
\caption{All pairwise win rates vs.\ the frozen Ideogram-4 base on the 1k Arena validation prompts, judged
by Gemini-3.5-flash and re-judged by GPT-5.4.}
\label{tab:ideo-ladder}
\begin{tabular}{@{}l cc cc@{}}
\toprule
& \multicolumn{2}{c}{Gemini-3.5-flash}
& \multicolumn{2}{c}{GPT-5.4} \\
\cmidrule(lr){2-3}\cmidrule(l){4-5}
Id & Arena RM & PickScore & Arena RM & PickScore \\
\midrule
\multicolumn{5}{@{}l}{\textit{Baseline}} \\
Arena-T2I-Hard
& .401 & .524 & .579 & 0.451 \\
preference only
& .266 & .334 & .285 & 0.403 \\
\midrule
\multicolumn{5}{@{}l}{\textit{Ours}} \\
Stage~1 (w/o anti-hack)
& .553 & .537 & .623 & .525 \\
\bottomrule
\end{tabular}
\end{table}

We further apply our Stage~1 recipe to Ideogram-4~\cite{ideogram2025ideogram}, an independently developed production model. We found that post-training can substantially degrade its text-rendering capability. We therefore include an additional OCR reward in all experiments reported in \cref{tab:ideo-ladder}. Details of the OCR reward are provided in \cref{sec:suppl-ocr}. \cref{tab:ideo-ladder} reproduces the same ordering observed on FLUX.2. Preference-only optimization performs substantially below parity, while Arena-T2I-Hard recovers much of the gap but still underperforms the frozen base, with a win rate of $0.4005$. Adding the intent-gated constraint reward is again the key step that pushes performance beyond parity, raising the win rate to $0.5526$. As we did not observe clear reward-hacking failure modes in the Stage~1 checkpoint, we do not introduce a Stage~2 and Stage~3 for Ideogram-4.

\subsection{Elo on the LMArena leaderboard}
\label{sec:leaderboard}
\paragraph{Settings} We test our Stage~1 and Stage~3 models in real-world large-scale live rating in Arena leaderboard. For FLUX.2-dev evaluation, we reproduce the FLUX.2-dev serving configuration and keep all inference settings fixed across the compared models. Specifically, we use 50 inference steps with CFG $4.0$, and generate images at a resolution of $1024\times1024$, matching the default FLUX.2-dev configuration on Hugging Face. For Ideogram-4, we use the same \emph{default\_quality\_48}. More details can be found in \cref{sec:suppl-serving}. All models are evaluated via live, blind, pairwise human comparisons against the production model pool on the Arena text-to-image leaderboard.

\paragraph{Results}
As shown in \cref{tab:ledger-leaderboard}, post-training establishes a new state of the art among open models on the live LMArena text-to-image leaderboard. Our strongest deployment, the Stage~1 post-trained Ideogram-4 model, achieves an Elo score of $1223.5$, surpassing every publicly listed open-source model; the previous best, Ideogram-4.0-Quality, trails it by $19.5$ Elo points. We observe similarly large gains on FLUX.2-dev: the Stage~1 model improves the frozen base from $1132.8$ to $1190.3$, while the Stage~3 further raises the score to $1202.1$, bringing FLUX.2-dev itself to the frontier of open-model performance. Together, these results show that our post-training recipe can generalize to live, blind human preference under production serving conditions, pushing open-model performance beyond the previous public state of the art.

%% file: sec/5_limitations.tex
\section{Limitations}
\label{sec:limitations}

Our study has several limitations. First, we evaluate our approach on only two base models, which limits our ability to assess how well the findings generalize across different model families and scales. Second, we do not evaluate our models in agentic workflows, where multi-step interactions and tool use may introduce additional challenges. Third, the decomposed faithfulness questions are not audited by human annotators, leaving open the possibility of errors or ambiguities in the decomposition process. Finally, due to computational constraints, we report results from a single random seed.

%% file: sec/6_conclusion.tex
\section{Conclusion}
\label{sec:conclusion}

We presented a simple and effective recipe for post-training open-domain text-to-image models, built on the composition of complementary reward signals: a Bradley--Terry preference reward trained on large-scale human votes, and open-ended, prompt-conditioned rubric rewards that captures prompt faithfulness, intent-conditional constraints and detectors for emerging hack modes. Furthermore, we introduce four mechanism to intergate the reward, making it robustness to optimization and generalize well. The resulting recipe reliably improves already-strong base models under live human evaluation: our RL-trained FLUX.2-dev gains $69$ Elo points over its base model on the LMArena text-to-image leaderboard, and our post-trained Ideogram-4 reaches an Elo of $1223.5$, surpassing every open-source model on the leaderboard. We believe this composition principle offers a scalable path for post-training future frontier generative models.

%% file: sec/X_suppl.tex
\section{Supplementary material}
\label{sec:suppl}

\subsection{Preference data and reward-model training details}
\label{sec:suppl-rmdetails}

\paragraph{Collator format} Each training example replicates one vote as
a chat-formatted sequence: a fixed reward-model system instruction, the
shared prompt as the user turn, and one of the two competing images as a
single assistant turn terminated by \texttt{<|cls|>}.

\paragraph{Optimization} We use 1 epoch with a global batch size of $1{,}032$ pairs, constant learning rate
$2\times10^{-6}$ with no warmup and set the max sequence ledngth to be $8{,}192$ tokens.

\paragraph{Augmentation values} Photometric augmentation
is applied to every training image inside the collator, after
aspect-ratio correction: brightness, contrast, and saturation jitter of
$\pm 0.05$ each, hue shift of $\pm 0.025$, and additive Gaussian pixel
noise at $\sigma{=}0.005$ on $[0,1]$-scaled pixels, applied to every training image.

\subsection{RL-settings details}
\label{sec:suppl-rlsettings}

\subsubsection{Training hyperparameters}
\label{sec:suppl-hparams}

\Cref{tab:suppl-hparams} lists the full rollout and optimization hyperparameters for both base models.

\begin{table}[htb]
  \centering
  \footnotesize
  \setlength{\tabcolsep}{4pt}
  \renewcommand{\arraystretch}{0.95}
  \caption{Training hyperparameters for the two base models.}
  \label{tab:suppl-hparams}
  \begin{tabular}{@{}lcc@{}}
    \toprule
     & FLUX.2 [dev] & Ideogram 4 \\
    \midrule
    \emph{Rollouts (per training epoch)} & & \\
    Prompt groups $G$ & 48 & 48 \\
    Rollouts per prompt $K$ & 24 & 24 \\
    Images per epoch & 1{,}152 & 1{,}152 \\
    Sampling steps & 10 & 10 \\
    Resolution & $512^2$ & $512^2$ \\
    Guidance weight & 4.0 & 7.0 \\
    Rollout solver & det.\ flow ODE & det.\ flow ODE \\
    NFT noise level & 0.7 & 0.7 \\
    \midrule
    \emph{LoRA (base weights frozen)} & & \\
    Rank / $\alpha$ & 64 / 128 & 32 / 64 \\
    Target modules & DiT only & DiT only \\
    Base-model precision & bf16 & fp8 \\
    LoRA precision & bf16 & bf 16  \\
    \midrule
    \emph{Optimization} & & \\
    Optimizer & AdamW & AdamW \\
    Adam $(\beta_1, \beta_2)$ & (0.9, 0.999) & (0.9, 0.999) \\
    Adam $\epsilon$ & $10^{-8}$ & $10^{-8}$ \\
    Learning rate & $3\times10^{-4}$ & $1.5\times10^{-4}$ \\
    Weight decay & $10^{-4}$ & $10^{-4}$ \\
    Max gradient norm & 1.0 & 1.0 \\
    NFT mixing coefficient $\beta$ & 1.0 & 1.0 \\
    KL regularization weight & $10^{-4}$ & $10^{-3}$ \\
    Trained timestep fraction & 0.99 & 0.99 \\
    EMA & yes & yes \\
    Micro-batch size per GPU & 3 & 3 \\
    Gradient-accumulation steps & 12 & 6 \\
    Optimizer updates per epoch & 1 & 1 \\
    Effective batch (images/update) & 1{,}152 & 1{,}152 \\
    GPUs & 32 & 64 \\
    Training epochs & 150 & 150 \\
    \bottomrule
  \end{tabular}
\end{table}

\subsubsection{Training-prompt filtering}
\label{sec:suppl-promptfilter}

\paragraph{Prompt-validator rubric.} The training-prompt filter is \texttt{gemini-3.5-flash} under a fixed yes/no rubric: a
prompt is kept only if it describes a single renderable image, is
predominantly visual description, depends on no external image or hidden
context, does not hinge on niche named-entity knowledge, is fluent
English, and carries enough visual detail to support a vivid image.

\paragraph{Validator system prompt.}
The system prompt given to \texttt{gemini-3.5-flash}
to validate candidate RL training prompts; a prompt
is kept only if the model answers \texttt{yes}.
\begin{lstlisting}[style=prompt]
You are an ultra-strict validator for text-to-image (T2I) prompts.

Goal: Decide whether the input string is a clean, single-image prompt that describes ONE concrete, visualizable subject/scene an artist could render WITHOUT any external image, slide context, or extra deliverables.

You MUST respond with ONLY:
yes   (VALID)
no    (INVALID)

No explanations. No extra text. No punctuation. No formatting changes.

--------------------------------
VALID ("yes") ONLY IF ALL ARE TRUE
1) Single-image describability
- The text directly describes one image (subject/scene/portrait/object/abstract artwork) that could be drawn as a single frame.

2) Visual-first content
- Most of the text is visual description (what is seen), optionally with style/medium/camera notes.

3) No dependency on hidden context
- The prompt does NOT reference any provided/previous image.

4) No strong external knowledge / named-entity dependency (ALLOW some well-known ones)
- INVALID if it depends on niche or obscure named entities as the *main* descriptor.
- VALID if any named entity used is widely recognizable OR the prompt still contains enough plain visual description to render the image without that knowledge.

5) Natural, fluent English
- The prompt must be written in natural, fluent English sentences or phrases.

6) Sufficient detail / not too short
- The prompt must contain enough visual detail to support a vivid image.
- INVALID if it is extremely short or generic with no extra descriptors.

--------------------------------
INVALID ("no") IF ANY ARE TRUE

A) Image editing or external-image dependency
- Mentions editing/adding to an existing image or references like:
"in the picture", "in this photo", "this same figurine", "characters in the picture", "as shown above/below", "provided image".

B) The prompt is too short to support a vivid image.

--------------------------------
Quick self-check (must be YES to return "yes"):
"Could an artist draw ONE specific image from this text alone, and is it primarily a visual description rather than a slide/marketing/logo/text-overlay request?"
\end{lstlisting}

\paragraph{Effect of prompt filtering.} We train Stage~1 model with Arena RM on both $100$k unfiltered arena prompts and our $10$k filtered arena prompts, and use the same judge protocol with GEMINI. The run trained on the unfiltered pool lost to its filtered counterpart by $0.0360$. We therefore read the gap as lower-quality prompts degrading training under the same optimization pressure, and adopt filtered prompt sets throughout the paper.

\subsection{Reward-checklist construction, statistics, and examples}
\label{sec:suppl-checklist-examples}

All checklists and intent labels are generated \emph{offline, once per
training prompt}, and stored alongside the prompt; the only model called
inside the RL loop is the yes/no checklist judge
served with vLLM.

\subsubsection{Checklist judge}
\label{sec:suppl-checklist-judge}

The VLM judge that answers checklist questions during
training. The batched form is used
whenever a dependency level contains several questions; the
single-question form is the fallback. All the rubric rewards share this judge; only
the question sets differ. Single-question form:
\begin{lstlisting}[style=prompt]
You are an impartial image quality judge. You will be given:
1. The original text-to-image prompt.
2. A specific yes/no verification question about the image.
3. The generated image.

Your task: look at the image and answer the question.

Rules:
- Answer with exactly one word: "yes", "no", or "irrelevant".
  - "yes"        = the image clearly satisfies the question.
  - "no"         = the image clearly does NOT satisfy the question.
  - "irrelevant" = the question does not apply to this image or cannot be determined from the image.
- Do NOT explain. Output ONLY one of the three words.
\end{lstlisting}
Batched form:
\begin{lstlisting}[style=prompt]
You are an impartial image quality judge. You will be given:
1. The original text-to-image prompt.
2. A list of yes/no verification questions about the image, each with an integer id.
3. The generated image.

Your task: look at the image and answer ALL questions in order.

Rules:
- For each question, answer "yes", "no", or "irrelevant".
  - "yes"        = the image clearly satisfies the question.
  - "no"         = the image clearly does NOT satisfy the question.
  - "irrelevant" = the question does not apply or cannot be determined.
- Output ONLY a JSON array of objects, each with "id" (int) and "answer" (string).
- Do NOT explain. No markdown fences. ONLY the JSON array.

Example output:
[{"id": 0, "answer": "yes"}, {"id": 1, "answer": "no"}, {"id": 2, "answer": "irrelevant"}]
\end{lstlisting}

\subsubsection{Faithfulness questions}
\label{sec:suppl-faithq}

A single decomposition call per prompt to \texttt{gemini-3-pro-preview} produces atomic yes/no questions with an explicit dependency DAG. The system prompt enforces: each question
checks exactly one visual attribute like object existence, color, count,
spatial relation, action, style, and text content; every detail of the
prompt is covered, ordered from core subjects to minor stylistic
details; nothing the prompt does not mention or imply may be asked
about; existence questions are DAG roots, attribute questions depend on
their object's existence question, and relation questions depend on
both objects' existence. The split carries $182{,}230$ faithfulness
questions  with its mean as $18.2$ per prompt, median as $17$, and max as $70$. At training
time the judge answers the DAG level-by-level in batched yes/no calls;
a question whose parent failed is scored ``no'' without being asked,
and the reward is the fraction of questions answered ``yes''.

\subsubsection{Constraint questions and the intent tag}
\label{sec:suppl-constrq}

A offline call per prompt to \texttt{gemini-3.1-pro-preview} does two things at once.
First it classifies the prompt's openness to unrequested additions:
\emph{strict} when the prompt closes the spec,
\emph{open} when a short or evocative prompt invites embellishment, and
\emph{neutral} in between; it also extracts a \texttt{style\_lock} indicating a
named style that must be preserved, a
\texttt{background\_spec}, and a coarse detail
budget. Second, it writes the constraint checklist strict prompts get one question per explicit negative,
a background check, a style-preservation check, and always the generic
catch-all \emph{``Is the image free of prominent objects, scenery, or
decorative details that the prompt did not request?''};
neutral prompts get only constraints explicitly stated in the prompt; open prompts get none. Label distribution: $3{,}140$ strict /
$4{,}443$ neutral / $2{,}405$ open / $12$ failure; $26{,}213$ constraint questions in
total. Constraint questions are phrased
so that ``yes'' means \emph{compliant} and are scored by the same yes/no judge as the faithfulness checklist.

\paragraph{Intent-gate examples.}
\emph{Strict}: ``top down pixel trader space ship with
no background''. \emph{Neutral}: ``please generate a scene of Seattle in the style of
Richard Scarry. Include the space needle, a ferry and pike place
market''. \emph{Open}: ``cat in red moccasins''.

\subsubsection{Graded OCR reward for Ideogram-4}
\label{sec:suppl-ocr}

The Ideogram-4 runs add a
graded OCR axis, targeting Ideogram's text-rendering emphasis. For each training prompt, the text spans
the prompt explicitly asks to be rendered like quoted strings and named
text elements are extracted once and stored as prompt metadata. During training, we use a VLM to
transcribe all visible text in the image \emph{without seeing the
prompt}, and each expected span receives
$0.5\cdot\mathrm{exact} + 0.3\cdot\mathrm{partial} + 0.2\cdot\mathrm{recall}$, where \emph{exact} indicates the span appearing
verbatim as a substring of the transcription, \emph{partial} is a fuzzy
string-similarity score, and \emph{recall} is the fraction of the
span's tokens recovered anywhere in
the transcription; the sample's OCR reward is the mean over its spans.
Prompts that request no text, and samples whose transcription call
fails, are excluded from the axis rather than penalized. The axis
enters with weight $1.0$ under the same per-batch z-normalization as
the other axes.

\subsubsection{Worked examples and hack-prevention detectors}
\label{sec:suppl-hackdetectors}

\begin{figure*}[p]
\centering
\scriptsize

\begin{minipage}[t]{0.195\textwidth}\vspace{0pt}\centering
\includegraphics[width=\linewidth]{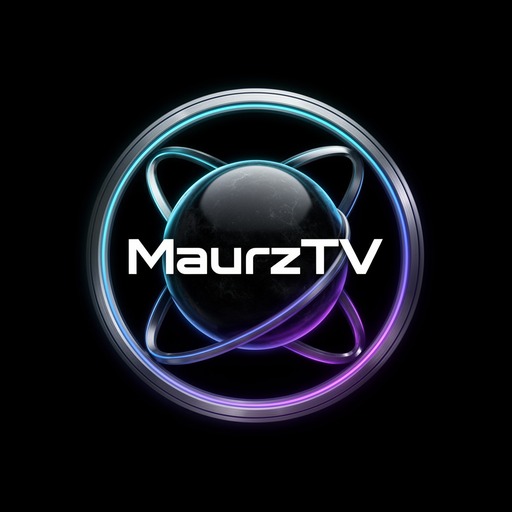}\\[1pt]
base (as served)\\[4pt]
\includegraphics[width=\linewidth]{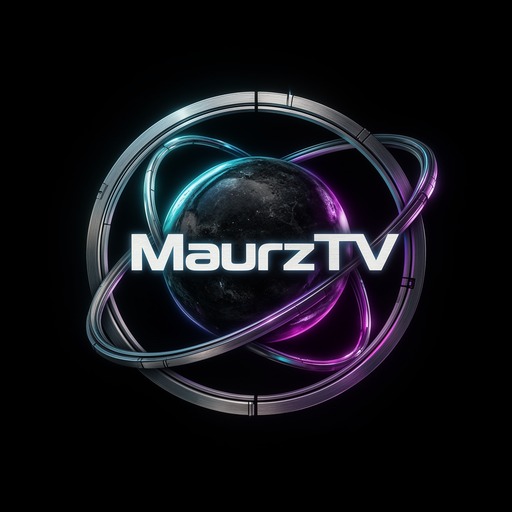}\\[1pt]
champion (as served)
\end{minipage}\hfill
\begin{minipage}[t]{0.775\textwidth}\vspace{0pt}\raggedright
\textbf{Example A.}
\emph{``logo, ultra modern, sophisticated, space and science, round for
Youtube, no background, name is MaurzTV, this is a channel with several
programs sharing the same topics or themes''}\\[1pt]
intent: \texttt{strict}\,; style lock: \texttt{logo}\,; background spec:
\texttt{no background}\,; detail budget: \texttt{minimal}\\[3pt]
\textbf{Faithfulness checklist} (presence; id\,[depends on]\,question):\\
F0\,[--]\; Is the image a logo?\\
F1\,[--]\; Is the text `MaurzTV' clearly visible in the image?\\
F2\,[0]\; Is the logo round or circular in shape?\\
F3\,[0]\; Does the logo feature space-related visual elements?\\
F4\,[0]\; Does the logo feature science-related visual elements?\\
F5\,[0]\; Does the logo have an ultra-modern design?\\
F6\,[0]\; Does the logo have a sophisticated appearance?\\
F7\,[--]\; Does the image have no background (transparent or completely plain)?\\[3pt]
\textbf{Constraint checklist} (absence; full strict battery, applied because the intent label is \texttt{strict}):\\
C0\; Does the image clearly feature the exact text `MaurzTV'?\\
C1\; Is the primary design of the logo round or circular?\\
C2\; Is the background completely plain, white, or transparent, with nothing else added?\\
C3\; Is the image rendered in a logo style, without drifting toward photorealistic or cinematic rendering?\\
C4\; Is the image free of prominent objects, scenery, or decorative details that the prompt did not request?
\end{minipage}\\[6pt]
\rule{\textwidth}{0.3pt}\\[6pt]

\begin{minipage}[t]{0.195\textwidth}\vspace{0pt}\centering
\includegraphics[width=\linewidth]{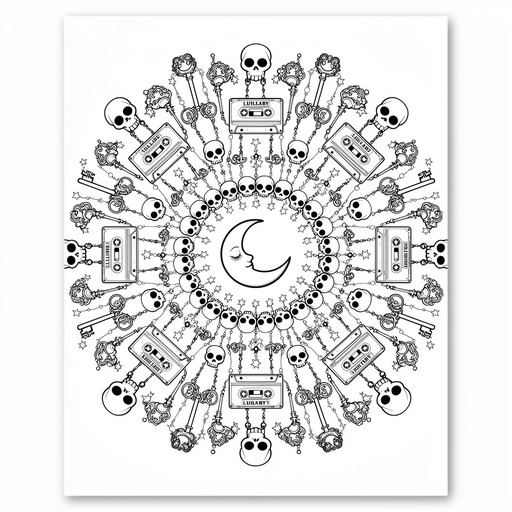}\\[1pt]
base (as served)\\[4pt]
\includegraphics[width=\linewidth]{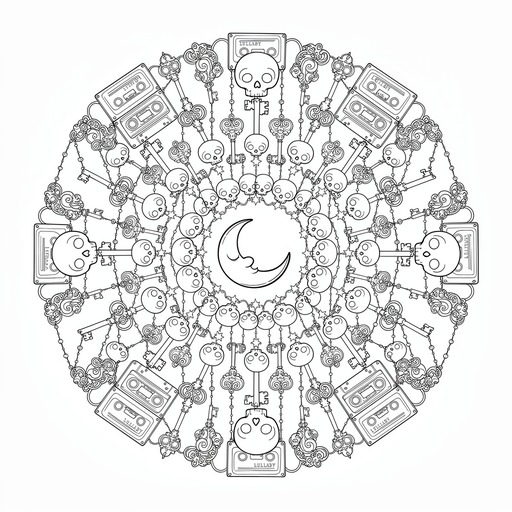}\\[1pt]
\end{minipage}\hfill
\begin{minipage}[t]{0.775\textwidth}\vspace{0pt}\raggedright
\textbf{Example B}
\emph{``Create an 8.5 in $\times$ 11 in portrait coloring book page, 300
DPI, 0.5 in margin: a symmetrical mandala made from lullaby mobile
charms (moons, tiny skulls, cassettes, keys) with repeating patterns for
coloring.''}\\[1pt]
intent: \texttt{strict}\,; style lock: \texttt{coloring book page}\,;
background spec: \texttt{plain white}\,; detail budget: \texttt{minimal}\\[3pt]
\textbf{Faithfulness checklist} (presence; id\,[depends on]\,question):\\[1pt]
\begin{minipage}[t]{0.485\linewidth}\raggedright
F0\,[--]\; Is there a mandala in the image?\\
F1\,[0]\; Is the mandala symmetrical?\\
F2\,[0]\; Does the mandala consist of repeating patterns?\\
F3\,[--]\; Are there moons in the image?\\
F4\,[0,3]\; Are the moons part of the mandala?\\
F5\,[--]\; Are there skulls in the image?\\
F6\,[5]\; Are the skulls tiny?\\
F7\,[0,5]\; Are the skulls part of the mandala?
\end{minipage}\hfill
\begin{minipage}[t]{0.485\linewidth}\raggedright
F8\,[--]\; Are there cassettes in the image?\\
F9\,[0,8]\; Are the cassettes part of the mandala?\\
F10\,[--]\; Are there keys in the image?\\
F11\,[0,10]\; Are the keys part of the mandala?\\
F12\,[0]\; Do the elements in the mandala resemble lullaby mobile charms?\\
F13\,[--]\; Is the image styled as a coloring book page (e.g., black and white line art)?\\
F14\,[--]\; Is the image in portrait orientation?\\
F15\,[--]\; Is there a visible margin around the artwork?
\end{minipage}\\[3pt]
\textbf{Constraint checklist} (absence; full strict battery):\\
C0\; Is the image a portrait-oriented symmetrical mandala composed of moons, tiny skulls, cassettes, and keys, with repeating patterns?\\
C1\; Does the image feature a visible blank margin around the edges?\\
C2\; Is the background plain white with nothing else added?\\
C3\; Is the image rendered in coloring book page style, without drifting toward photorealistic or cinematic rendering?\\
C4\; Is the image free of prominent objects, scenery, or decorative details that the prompt did not request?
\end{minipage}\\[6pt]
\rule{\textwidth}{0.3pt}\\[6pt]

\begin{minipage}[t]{\textwidth}\raggedright
\textbf{Hack-prevention detector questions};
fixed per failure mode, not generated per prompt; verdicts enter the
objective only as one-sided vetoes, never as reward:\\[2pt]
\emph{Text gibberish} (pointwise; armed on every sample):
``Is the image free of garbled, misspelled, or meaningless rendered text
(random letters, gibberish words, keyboard rows) and free of the same
text being unintentionally repeated many times?''\\[2pt]
\emph{Photo-style drift} (reference-based; armed on both prompts above
because each locks a non-photographic style and a frozen-base reference
render exists; judged in both presentation orders):
``Which image is rendered in a more photorealistic style - more
realistic textures, photographic lighting, camera-like detail? The
first image is Image A, the second is Image B. Answer A, B, or same.''
\end{minipage}
\caption{The complete per-prompt reward battery for two real evaluation prompts, shown alongside the as-served renders from the FLUX.2 base model and the Stage~1 model. For each prompt, we list every question posed by the reward: the presence-oriented faithfulness DAG, in which a question whose parent fails is automatically scored `no'' without being asked; the absence-oriented constraint checklist, generated in full here because both prompts are labeled \texttt{strict}; and, at the bottom, the two armed hack-prevention detector questions shared across prompts.}

\label{fig:suppl-qa-examples}
\end{figure*}

In \cref{fig:suppl-qa-examples}, we show everything the reward sees for two real
evaluation prompts: the
as-served base and champion renders, the complete faithfulness DAG, the
complete constraint checklist with its intent header, and the verbatim
hack-prevention detector questions with their arming conditions.

\subsubsection{Human audit of the intent labels}
\label{sec:suppl-gateaudit}

The strict/neutral/open label is one automated classifier call per
prompt, so we audited it against human judgment on a $400$-prompt
subset of the evaluation prompts. Two
authors labeled every prompt from the original user text.

\begin{center}
\begin{tabular}{@{}lcccc@{}}
\toprule
 & \multicolumn{3}{c}{classifier} & \\
\cmidrule(lr){2-4}
human & strict & neutral & open & $\Sigma$ \\
\midrule
strict  & \textbf{100} & 21  & 1  & 122 \\
neutral & 12  & \textbf{128} & 36 & 176 \\
open    & 8   & 11  & \textbf{83} & 102 \\
\midrule
$\Sigma$ & 120 & 160 & 120 & 400 \\
\bottomrule
\end{tabular}
\end{center}

The adjudicated result is $311/400$ agreement with an accuracy at $77.8\%$ and $\kappa=0.66$. What the gate actually consumes is
the binary strict-versus-not decision, and there human and classifier
agree on $89.5\%$.

\subsection{Evaluation details}
\label{sec:suppl-eval}

\subsubsection{Pairwise judging protocol}
\label{sec:suppl-judgeprotocol}

Existing benchmarks and reward models typically emphasize only one or
a few dimensions of generation quality, and therefore may not fully
capture overall user preference in real-world settings; this motivates
the pairwise LLM-as-a-judge protocol. For each
prompt the judge receives the two generated images, the original user
prompt, and a fixed evaluation rubric, and selects the better
generation; to reduce position bias, each image pair is evaluated twice
with the presentation order reversed and the verdicts aggregated.
The canonical FLUX.2 protocol renders the 1k Arena validation
prompts. Both the evaluated
checkpoint and the frozen base generate from prompts expanded at
inference time by \texttt{gemini-3.5-flash} with FLUX.2's and Ideogram 4's official message. The judge
receives the original, unrewritten user prompt on both sides.
\paragraph{Judging rubric.} We use MMRB2's image-generation judge
prompt verbatim. The judge receives the original, unrewritten user
prompt and the two images as Response A / Response B, reasons over the
rubric's seven criteria (\texttt{faithfulness\_to\_prompt},
\texttt{text\_rendering}, \texttt{input\_faithfulness},
\texttt{image\_consistency}, \texttt{text\_image\_alignment},
\texttt{text\_quality}, \texttt{overall\_quality} and returns structured JSON: a $1$--$6$ comparative score, a
\texttt{better\_response} of A or B, per-criterion winners, and a
confidence value. The overall verdict is read from
\texttt{better\_response}; the faithfulness and aesthetic subscores
are read from the \texttt{faithfulness\_to\_prompt} and
\texttt{overall\_quality} criterion winners. The input-image
and multi-image criteria are ``Not Applicable'' for single-image
T2I,

\paragraph{MMRBv2 pairwise evaluation judge.}
The MMRBv2 image-generation judge prompt.
\begin{lstlisting}[style=prompt]
You are an expert in multimodal quality analysis and generative AI evaluation. Your role is to act as an objective judge for comparing two AI-generated responses to the same prompt. You will evaluate which response is better based on a comprehensive rubric.

**Important Guidelines:**
- Be completely impartial and avoid any position biases
- Ensure that the order in which the responses were presented does not influence your decision
- Do not allow the length of the responses to influence your evaluation
- Do not favor certain model names or types
- Be as objective as possible in your assessment
- Consider factors such as helpfulness, relevance, accuracy, depth, creativity, and level of detail

**Understanding the Content Structure:**
- **[ORIGINAL PROMPT TO MODEL:]**: This is the instruction given to both AI models
- **[INPUT IMAGE FROM PROMPT:]**: This is the source image provided to both models (if any)
- **[RESPONSE A:]**: The first model's generated response (text and/or images)
- **[RESPONSE B:]**: The second model's generated response (text and/or images)

Your evaluation must be based on a fine-grained rubric that covers the following criteria. For each criterion, you must provide detailed step-by-step reasoning comparing both responses. You will use a 1-6 scoring scale.

**Evaluation Criteria:**
1. **faithfulness_to_prompt:** Which response better adheres to the composition, objects, attributes, and spatial relationships described in the text prompt?

2. **text_rendering:** If either response contains rendered text, which one has better text quality (spelling, legibility, integration)? If no text is rendered, state "Not Applicable."

3. **input_faithfulness:** If an input image is provided, which response better respects and incorporates the key elements and style of that source image? If no input image is provided, state "Not Applicable."

4. **image_consistency:** If multiple images are generated, which response has better visual consistency between images (character appearance, scene details)? If no multiple images are provided, state "Not Applicable."

5. **text_image_alignment:** Which response has better alignment between text descriptions and visual content?

6. **text_quality:** If text was generated, which response has better linguistic quality (correctness, coherence, grammar, tone)?

7. **overall_quality:** Which response has better general technical and aesthetic quality, realism, coherence, and fewer visual artifacts or distortions?

**Scoring Rubric:**
- Score 6 (A is significantly better): Response A is significantly superior across most criteria
- Score 5 (A is marginally better): Response A is noticeably better across several criteria
- Score 4 (Unsure or A is negligibly better): Response A is slightly better or roughly equivalent
- Score 3 (Unsure or B is negligibly better): Response B is slightly better or roughly equivalent
- Score 2 (B is marginally better): Response B is noticeably better across several criteria
- Score 1 (B is significantly better): Response B is significantly superior across most criteria

**Confidence Assessment:**
After your evaluation, assess your confidence in this judgment on a scale of 0.0 to 1.0:

**CRITICAL**: Be EXTREMELY conservative with confidence scores. Most comparisons should be in the 0.2-0.5 range.

- **Very High Confidence (0.8-1.0)**: ONLY for absolutely obvious cases where one response is dramatically better across ALL criteria with zero ambiguity. Use this extremely rarely (less than 10% of cases).
- **High Confidence (0.6-0.7)**: Clear differences but some uncertainty remains. Use sparingly (less than 20% of cases).
- **Medium Confidence (0.4-0.5)**: Noticeable differences but significant uncertainty. This should be your DEFAULT range.
- **Low Confidence (0.2-0.3)**: Very close comparison, difficult to distinguish. Responses are roughly equivalent or have conflicting strengths.
- **Very Low Confidence (0.0-0.1)**: Essentially indistinguishable responses or major conflicting strengths.

**IMPORTANT GUIDELINES**:
- DEFAULT to 0.3-0.5 range for most comparisons
- Only use 0.6+ when you are absolutely certain
- Consider: Could reasonable people disagree on this comparison?
- Consider: Are there any strengths in the "worse" response?
- Consider: How obvious would this be to a human evaluator?
- Remember: Quality assessment is inherently subjective

After your reasoning, you will provide a final numerical score, indicate which response is better, and assess your confidence. You must always output your response in the following structured JSON format:

{
    "reasoning": {
        "faithfulness_to_prompt": "YOUR REASONING HERE",
        "text_rendering": "YOUR REASONING HERE",
        "input_faithfulness": "YOUR REASONING HERE",
        "image_consistency": "YOUR REASONING HERE",
        "text_image_alignment": "YOUR REASONING HERE",
        "text_quality": "YOUR REASONING HERE",
        "overall_quality": "YOUR REASONING HERE",
        "comparison_summary": "YOUR OVERALL COMPARISON SUMMARY HERE"
    },
    "score": <int 1-6>,
    "better_response": "A" or "B",
    "confidence": <float 0.0-1.0>,
    "confidence_rationale": "YOUR CONFIDENCE ASSESSMENT REASONING HERE"
}
\end{lstlisting}

\paragraph{Forward/reverse aggregation.} Each image pair is judged
twice at temperature $0$ with the presentation order swapped, and each ordering counts as one unit. The reported win
rate is the candidate's units over all $2n$ orderings: a pair the
candidate wins in both orders contributes $1$, a clean loss $0$, and
a position-flip disagreement exactly $1/2$. The faithfulness and
aesthetic subscores aggregate the per-criterion winners the same way.

\subsubsection{Ideogram-4 serving configuration}
\label{sec:suppl-serving}

All Ideogram-4 numbers use the production
\texttt{V4\_QUALITY\_48} sampling configuration released in the official repo\footnote{https://github.com/ideogram-oss/ideogram4.git}.

\subsection{Prompt-rewrite system message}
\label{sec:suppl-systemprompt}

We reuse FLUX.2/Ideogram-4's own official
upsampling system
message~\cite{blackforestlabs2024flux,blackforestlabs2026flux2, ideogram2025ideogram} verbatim as
the system instruction to \texttt{gemini-3.5-flash}.

\paragraph{FLUX.2.} The official FLUX.2 upsampling system message below is used verbatim with the sampling temperature as $0.3$; the rewriter returns only the expanded prompt.

\begin{lstlisting}[style=prompt]
You are an expert prompt engineer for FLUX.2 by Black Forest Labs. Rewrite user prompts to be more descriptive while strictly preserving their core subject and intent.

Guidelines:
1. Structure: Keep structured inputs structured (enhance within fields). Convert natural language to detailed paragraphs.
2. Details: Add concrete visual specifics - form, scale, textures, materials, lighting (quality, direction, color), shadows, spatial relationships, and environmental context.
3. Text in Images: Put ALL text in quotation marks, matching the prompt's language. Always provide explicit quoted text for objects that would contain text in reality (signs, labels, screens, etc.) - without it, the model generates gibberish.

Output only the revised prompt and nothing else.
\end{lstlisting}

\paragraph{Ideogram 4.} Ideogram's magic-prompt v1 system message is a $\sim$4{,}500-word structured-captioning contract that ships verbatim in the \texttt{magic\_prompt\_system\_prompts/v1.txt} in model's open-weight release; we apply it unmodified. We reproduce its opening output contract below and refer to the released file for the full text.

\begin{lstlisting}[style=prompt]
[SYSTEM]
You convert a natural-language user idea into a structured JSON caption an image renderer can consume. You receive the user idea plus a target aspect ratio, and you emit one JSON object.

## OUTPUT CONTRACT - exactly three top-level keys, in this order:

{"aspect_ratio":"W:H","high_level_description":"...","compositional_deconstruction":{"background":"...","elements":[ ... ]}}

- Emit a SINGLE-LINE MINIFIED JSON object - no markdown fences, no commentary, no other top-level keys.
- Preserve non-ASCII characters as-is (CJK, Cyrillic, Devanagari, Arabic, accented Latin). [...]

### `high_level_description` - observational summary (50-word hard cap)

- ONE long sentence preferred, never more than two.
- Reads like a short natural-language prompt, not an analysis. Starts immediately with the subject [...]

[... full ~4,500-word message continues in the released file ...]
\end{lstlisting}

\subsection{Weight-soup merge recipes}
\label{sec:suppl-soups}

All soups reported in \cref{tab:all-results,tab:ablation} are constructed directly from the LoRA adapters. Every run uses the same adapter configuration: rank $64$, $\alpha=128$, dropout $0$, and adapters on $144$ modules of the FLUX.2 transformer. All merges use fixed equal weights with$0.5/0.5$.

\paragraph{Arena cross-run soup}
The Arena-RM \emph{cross-soup} combines best checkpoints of the gated-constraint run and the veto run by averaging their induced LoRA weight updates,

$$
\Delta W_{\mathrm{cross}}
=
\tfrac12\Delta W_{\mathrm{champ}}
+
\tfrac12\Delta W_{\mathrm{veto}}.
$$

We realize this average exactly through rank concatenation. Writing a LoRA update as
$\Delta W=(\alpha/r)BA$, we construct

$$
A_{\mathrm{cross}}
=
\begin{bmatrix}
A_{\mathrm{champ}}\\
A_{\mathrm{veto}}
\end{bmatrix},
\qquad
B_{\mathrm{cross}}
=
\begin{bmatrix}
\tfrac12 B_{\mathrm{champ}} &
\tfrac12 B_{\mathrm{veto}}
\end{bmatrix}.
$$

The rank therefore increases from $64$ to $128$, while $\alpha$ is scaled from $128$ to $256$ so that $\alpha/r$ remains unchanged. This construction preserves

$$
\Delta W_{\mathrm{cross}}
=
\tfrac12\Delta W_{\mathrm{champ}}
+
\tfrac12\Delta W_{\mathrm{veto}}
$$

exactly, without introducing cross terms between the two LoRA factorizations.

\paragraph{Same-run soups}
The Arena \emph{same-soup} averages checkpoints $60$ and $90$ from the Stage~2 trajectory directly in LoRA-factor space:

$$
A_{\mathrm{same}}
=
\tfrac12(A_{60}+A_{90}),
\qquad
B_{\mathrm{same}}
=
\tfrac12(B_{60}+B_{90}),
$$

with rank and $\alpha$ unchanged. Unlike the cross-run construction above, this operation is not an exact average in $\Delta W$ space:

$$
B_{\mathrm{same}}A_{\mathrm{same}}
=
\tfrac14
\left(
B_{60}A_{60}
+
B_{60}A_{90}
+
B_{90}A_{60}
+
B_{90}A_{90}
\right).
$$

It therefore contains cross terms between the two factorizations. The PickScore \emph{same-soup} control is constructed identically from checkpoints $60$ and $90$ of the PickScore-Stage~2 run.

\paragraph{PickScore cross-run soup}
The PickScore \emph{cross-soup} uses the same factor-wise equal average of the two LoRA adapters rather than the exact $\Delta W$ construction above. We retain this recipe as a control for the merge operation itself. Thus, the cross-run improvement reported in \cref{sec:m-crosssoup} is not specific to exact $\Delta W$ concatenation: a corresponding gain also appears under the simpler factor-wise merge.

\subsection{Transferability to an external evaluation prompt set}
\label{sec:m-mmrb2rw}

We further evaluate our reward design on the 791 public MMRBv2 T2I test prompts~\citep{hu2026multimodal}, using the same checkpoint selected on the arena validation set and prompt rewriting. None of these prompts are used for training, reward tuning, or checkpoint selection. As shown in \cref{tab:ledger-robustness}, the main trends observed on our primary evaluation set transfer to this external benchmark. Moreover, Stage~3 again achieves the highest win rate at $0.6011$, and the ablation orderings are preserved. The relative gains are smaller on MMRBv2, likely because its prompts are generally simpler and place less demand on the complex faithfulness and constraint satisfaction that real user prompts require. The table indicates that the gains from combining complementary reward designs generalize beyond the prompt distribution used during development.

\begin{table}[htb]
  \centering
  \footnotesize
  \setlength{\tabcolsep}{5pt}
  \renewcommand{\arraystretch}{0.92}
  \caption{Prompt-set transfer: we swap the prompt set
  to the 791 public MMRBv2 T2I test prompts. The checkpoints are
  selected on the 1k Arena validation prompts.}
  \label{tab:ledger-robustness}
  \begin{tabular}{@{}lc@{}}
    \toprule
    Id & MMRBv2 prompts \\
    \midrule
    \multicolumn{2}{@{}l}{\textit{Baseline}} \\
    faithfulness only & 0.3357 \\
    Arena-T2I-Hard & 0.5563 \\
    preference only & 0.5272 \\
    \midrule
    \multicolumn{2}{@{}l}{\textit{Ours}} \\
    Stage~1 & 0.5632 \\
    Stage~2 & 0.5632 \\
    Stage~3 & \textbf{0.6011} \\
    \midrule
    \multicolumn{2}{@{}l}{\textit{Ablation variants}} \\
    ungated constraint & 0.4987 \\
    additive hack detector & 0.4671 \\
    same-recipe soup & 0.5588 \\
    \bottomrule
  \end{tabular}
\end{table}

\subsection{Confidence intervals for the main tables}
\label{sec:suppl-ledger-cis}

\cref{tab:ledger-cis} reports percentile-bootstrap $95\%$
confidence intervals for every cell of \cref{tab:all-results}, and
\cref{tab:ledger-cis-ablation} for every cell of \cref{tab:ablation}. We use $10{,}000$ bootstrap replicates over prompt pairs with a fixed seed.

\begin{table}[htb]
\centering
\scriptsize
\setlength{\tabcolsep}{0.6pt}
\renewcommand{\arraystretch}{1.02}
\caption{$95\%$ percentile-bootstrap confidence intervals for
every cell of \cref{tab:all-results}. Each cell re-states the win rate vs.\
the frozen FLUX.2 base on the 1k Arena validation rewrite prompts together
with its interval. The DiffusionNFT and faithfulness-only baselines train
with neither preference reward, so their Arena-RM and PickScore columns
repeat the same run. All intervals use $10{,}000$ bootstrap replicates over
prompt pairs with a fixed seed.}
\label{tab:ledger-cis}
\begin{tabular}{@{}lcccc@{}}
\toprule
& \multicolumn{2}{c}{Gemini-3.5-flash} & \multicolumn{2}{c}{GPT-5.4} \\
\cmidrule(lr){2-3}\cmidrule(l){4-5}
Id & Arena RM & PickScore & Arena RM & PickScore \\
\midrule
\multicolumn{5}{@{}l}{\textit{Baseline}} \\
DiffusionNFT      & .290 [.266,\,.315] & .290 [.266,\,.315] & .396 [.370,\,.421] & .396 [.370,\,.421] \\
faithfulness only & .366 [.340,\,.392] & .366 [.340,\,.392] & .367 [.344,\,.390] & .367 [.344,\,.390] \\
Arena-T2I-Hard    & .589 [.561,\,.615] & .577 [.550,\,.603] & .557 [.533,\,.580] & .537 [.513,\,.560] \\
preference only   & .507 [.480,\,.533] & .461 [.434,\,.487] & .417 [.394,\,.441] & .517 [.492,\,.541] \\
\midrule
\multicolumn{5}{@{}l}{\textit{Ours}} \\
Stage~1           & .642 [.617,\,.666] & .581 [.554,\,.607] & .561 [.539,\,.584] & .611 [.587,\,.634] \\
Stage~2           & .635 [.609,\,.660] & .571 [.545,\,.597] & .567 [.544,\,.589] & .585 [.562,\,.607] \\
Stage~3           & .660 [.635,\,.683] & .626 [.600,\,.651] & .572 [.549,\,.595] & .623 [.601,\,.644] \\
\bottomrule
\end{tabular}
\end{table}

\begin{table}[htb]
\centering
\scriptsize
\setlength{\tabcolsep}{0.6pt}
\renewcommand{\arraystretch}{1.02}
\caption{$95\%$ percentile-bootstrap confidence intervals for
every cell of \cref{tab:ablation}. The intent-gated variant is the Stage~1
run, and the applicable-gated and one-sided-veto variants are the Stage~2
run, so those rows repeat the corresponding intervals of
\cref{tab:ledger-cis}. All intervals use $10{,}000$ bootstrap replicates
over prompt pairs with a fixed seed.}
\label{tab:ledger-cis-ablation}
\begin{tabular}{@{}lcccc@{}}
\toprule
& \multicolumn{2}{c}{Gemini-3.5-flash} & \multicolumn{2}{c}{GPT-5.4} \\
\cmidrule(lr){2-3}\cmidrule(l){4-5}
Variant & Arena RM & PickScore & Arena RM & PickScore \\
\midrule
\multicolumn{5}{@{}l}{\textit{Constraint gate}} \\
ungated          & .605 [.580,\,.630] & .520 [.493,\,.546] & .555 [.532,\,.579] & .498 [.475,\,.521] \\
intent-gated     & .642 [.617,\,.666] & .581 [.554,\,.607] & .561 [.539,\,.584] & .611 [.587,\,.634] \\
\midrule
\multicolumn{5}{@{}l}{\textit{Hack applicability gate}} \\
ungated          & .559 [.533,\,.586] & .570 [.543,\,.596] & .500 [.477,\,.522] & .551 [.529,\,.574] \\
applicable-gated & .635 [.609,\,.660] & .571 [.545,\,.597] & .567 [.544,\,.589] & .585 [.562,\,.607] \\
\midrule
\multicolumn{5}{@{}l}{\textit{Hack signal integration}} \\
additive         & .526 [.499,\,.553] & .419 [.393,\,.445] & .473 [.451,\,.495] & .410 [.389,\,.433] \\
one-sided veto   & .635 [.609,\,.660] & .571 [.545,\,.597] & .567 [.544,\,.589] & .585 [.562,\,.607] \\
\midrule
\multicolumn{5}{@{}l}{\textit{Weight ensembling}} \\
same-recipe      & .644 [.618,\,.668] & .561 [.535,\,.587] & .546 [.524,\,.569] & .568 [.545,\,.590] \\
cross-recipe     & .660 [.635,\,.683] & .626 [.600,\,.651] & .572 [.549,\,.595] & .623 [.601,\,.644] \\
\bottomrule
\end{tabular}
\end{table}

\subsection{Optimizing a single preference reward hurts faithfulness}
\label{sec:goodhart}
\begin{figure}[htb]
  \centering
  \includegraphics[width=\linewidth]{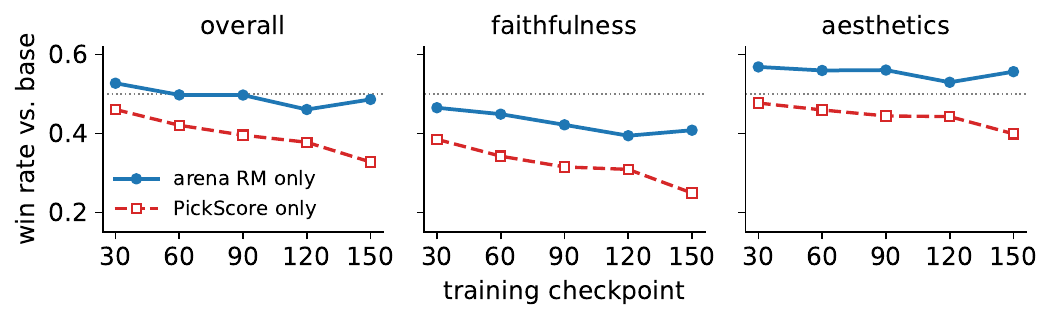}
  \caption{Single-preference training vs.\ base across checkpoints,
    evaluated on the $1$k arena validation prompts with
    \texttt{gemini-3.5-flash} rewrites. The faithfulness panel shows both independently trained preference rewards below parity and degrading with
    training, even while aesthetics holds at or above parity for the
    arena-RM run: faithfulness is traded away for aesthetics.}
  \label{fig:goodhart-perdim}
\end{figure}

We train models using a single preference reward and track their win rates against the base model across checkpoints and evaluation dimensions. As shown in \cref{fig:goodhart-perdim}, neither reward yields a reliable overall improvement: optimizing the arena reward reaches a best win rate of only $0.5065$, while PickScore remains below parity from the first evaluated checkpoint. More importantly, their faithfulness win rates fall below $50\%$ and continue to degrade with training.

We attribute this trade-off to the use of a single holistic preference score. Because the reward aggregates multiple aspects of image quality into one scalar, gains in dimensions that are easier to optimize, such as aesthetics, can compensate for losses in prompt faithfulness. The policy can therefore increase predicted preference while progressively omitting requested content.

